\documentclass[10pt,twocolumn,letterpaper]{article}
\usepackage{iccv}
\usepackage{times}
\usepackage{epsfig}
\usepackage{graphicx}
\usepackage{amsmath}
\usepackage{amssymb}
\usepackage{booktabs}
\usepackage{multirow}
\usepackage{threeparttable}
\usepackage[ruled]{algorithm2e}
\SetArgSty{textnormal}
\usepackage{array}
\usepackage{colortbl}
\usepackage[accsupp]{axessibility}
\usepackage[pagebackref=true,breaklinks=true,letterpaper=true,colorlinks,bookmarks=false]{hyperref}

\iccvfinalcopy 

\def\iccvPaperID{8} 

\ificcvfinal\fi

\begin{document}

\title{Improving Self-supervised Learning with Hardness-aware Dynamic Curriculum Learning: An Application to Digital Pathology}

\author{Chetan L Srinidhi, Anne L Martel \\
Physical Sciences, Sunnybrook Research Institute, Toronto, Canada \\
Department of Medical Biophysics, University of Toronto, Canada \\
\{\tt\small chetan.srinidhi, a.martel\}@utoronto.ca
}

\maketitle
\ificcvfinal\thispagestyle{empty}\fi

\begin{abstract}
Self-supervised learning (SSL) has recently shown tremendous potential to learn generic visual representations useful for many image analysis tasks. Despite their notable success, the existing SSL methods fail to generalize to downstream tasks when the number of labeled training instances is small or if the domain shift between the transfer domains is significant. In this paper, we attempt to improve self-supervised pretrained representations through the lens of curriculum learning by proposing a hardness-aware dynamic curriculum learning (\textbf{HaDCL}) approach. To improve the robustness and generalizability of SSL, we dynamically leverage progressive harder examples via \textbf{easy-to-hard} and \textbf{hard-to-very-hard} samples during mini-batch downstream fine-tuning. We discover that by progressive stage-wise curriculum learning, the pretrained representations are significantly enhanced and adaptable to both \textbf{in-domain} and \textbf{out-of-domain} distribution data. 

We performed extensive validation on \textbf{three} histology benchmark datasets on both \textbf{patch-wise} and \textbf{slide-level} classification problems. Our curriculum based fine-tuning yields a significant improvement over standard fine-tuning, with a minimum improvement in area-under-the-curve (AUC) score of 1.7\% and 2.2\% on in-domain and out-of-domain distribution data, respectively. Further, we empirically show that our approach is more generic and adaptable to any SSL methods and does not impose any additional overhead complexity. Besides, we also outline the role of patch-based versus slide-based curriculum learning in histopathology to provide practical insights into the success of curriculum based fine-tuning of SSL methods.\footnote{\textcolor{blue}{Code is released at \url{https://github.com/srinidhiPY/ICCV-CDPATH2021-ID-8}}} 
\end{abstract}
\vspace{-5mm}

\section{Introduction}
\label{sec:Introduction}
Learning with limited human supervision is a longstanding goal in machine learning, especially in medical image analysis due to the expensive and time-consuming annotation process. Self-supervised learning (SSL) methods have gained increasing popularity due to their ability to learn general-purpose features that are competitive with representations generated by state-of-the-art (SoTA) fully-supervised methods \cite{azizi2021big, chen2020simple, ciga2020self, srinidhi2021self}. These methods involve two steps: unsupervised pretraining on unlabeled data in a task-agnostic way, followed by supervised fine-tuning in a task-specific way with limited labeled data. SSL methods, however, often struggle to perform well on downstream tasks and generalize poorly on out-of-distribution data due to limited downstream supervision \cite{andreassen2021evolution, yan2020clusterfit, zhai2019s4l}. 

Recent studies have focused on improving self-supervised pretrained representations with effective sampling strategies that mine informative hard examples via aggressive data augmentations \cite{li2020self} or with hard-negative mining techniques \cite{robinson2020contrastive}. However, these methods are tailored to improve a specific family of contrastive based SSL methods (such as MoCo \cite{he2020momentum}, and SimCLR \cite{chen2020simple}) and cannot be applied or generalized to other pretraining methods. In contrast, consistency based semi-supervised techniques \cite{chen2020big, srinidhi2021self, zhai2019s4l} have been proposed to improve the SSL by utilizing the unlabeled data in a task-specific semi-supervised manner. However, despite their improved performance, the semi-supervised approaches typically suffer from the problem of confirmation bias \cite{arazo2020pseudo}, and as yet, their practical applicability to medical image analysis has been severely limited. 

In this paper, we attempt to improve self-supervised pretrained representations through the lens of curriculum learning (CL). CL in machine learning paradigm \cite{bengio2009curriculum, soviany2021curriculum} is fundamentally inspired by the human learning process, where the easier concepts (examples) are presented first, and most difficult concepts are learned later on. Such meaningful ordering of samples (as opposed to random ordering) during training has shown to improve both convergence speed and accuracy of the neural network model. To this end, we extend the previous idea of leveraging hard examples to improve the self-supervised pretrained representations \cite{li2020self, robinson2020contrastive} further by combining SSL and CL in an elegant manner. In this study, we empirically investigate their inter-dependencies and present novel ways to combine them to achieve faster convergence, better generalization ability, and alleviate over-fitting to in-domain data. In relation to the CL paradigm introduced in \cite{bengio2009curriculum, hacohen2019power, wu2020curricula}, we first start by asking two very fundamental questions: (i) \textbf{how to determine the notion of example difficulty} (i.e., \textit{scoring} function) that is made available to the network during training? (ii) \textbf{how to specify the order} (typically, easy to hard) \textbf{at which the examples are presented to the network?} - which depends on both the data and learning model.

To answer the above questions, we first attempt to determine the ``\textit{\textbf{hardness}}" or ``\textit{\textbf{difficulty}}" of each sample in the training data via curriculum by transfer learning approach, initially proposed in Weinshall \textit{et al.}  \cite{weinshall2018curriculum}. Here, we choose to rank the difficulty of training samples with the help of instantaneous feedback (i.e., \textit{loss}) from the pretrained self-supervised model, while fine-tuning on the downstream task of interest. Unlike in the previous study in histopathology \cite{wei2021learn}, where the ranking (i.e., hardness) of samples are determined with the aid of human teachers (i.e., pathologists), our proposed approach rather investigates the knowledge transfer to determine the hardness of each training sample, which provides more reliable scores for the target task and does not involve any additional human-intervention. This is particularly important in pathology, where obtaining multiple annotator agreements as a proxy for determining the sample difficulty is often time-consuming and challenging. Besides, it is also shown in previous studies \cite{hacohen2019power, weinshall2018curriculum} that the ranking provided by the human teachers may not reflect the true underlying difficulty as it affects the neural network. 

Second, we focus our attention on specifying the \textit{``\textbf{order}"} at which the data is presented to the network. Typically, most studies in the CL literature \cite{soviany2021curriculum, wu2020curricula} either follow ordering of input examples from easy-to-hard (curriculum) or hard-to-easy (anti-curriculum) and sometimes random \cite{wu2020curricula}. However, one significant limitation with the existing approaches is that they do not necessarily consider the learning dynamics of the neural network while estimating the sample hardness over the course of model training. Due to the stochastic mini-batch style nature of gradient-descent optimization, the instantaneous hardness of each training sample changes over time from the early part of training to the later part of training. i.e., the hardness of each sample decreases monotonically over the course of training, where the hard samples become easier, while easy samples stay easy throughout training. With this motivation, we choose to measure the sample hardness adaptively in each mini-batch during task-specific fine-tuning of self-supervised pretrained model by introducing ``hardness-aware dynamic curriculum learning (HaDCL)" as a mini-batch instantaneous hardness measure of a training sample over time. Empirically, we show that our proposed HaDCL strategy significantly improves the SSL on a challenging downstream task, i.e., breast cancer lymph node metastases detection on both in-domain and out-of-domain data, supporting that the hard example mining is indeed crucial for improved model accuracy and generalizability.  
\vspace{1.5mm}

\noindent
\textbf{Contributions.} To summarise, we make the following contributions in this study: 
\vspace{-2mm}
\begin{itemize}
    \item We propose a principled way of combining SSL with CL to improve the self-supervised pretrained representations on the downstream task on both in-domain and out-of-domain distribution data.
    \vspace{-2mm}
    \item We present a mini-batch hardness-aware dynamic curriculum learning (HaDCL) strategy to determine the instantaneous hardness of training samples with improved training convergence and better accuracy.
    \vspace{-2mm}
    \item We also conduct an empirical study to understand the boundaries within which the curriculum works to improve SSL on both patch-wise and slide-level classification tasks in histology. Further, we also probe the generalizability of our method on out-of-distribution data with significant domain shifts. 
\end{itemize}

\vspace{-2mm}
\section{Related Work}
\label{sec:Realted work}

\noindent
\textbf{Self-supervised Learning.}
Inspired by the recent success in SSL, the existing methods are categorized into context-based and contrastive-based learning methods. The early works focused on context-based methods to formulate an auxiliary task (i.e., \textit{pretext} task) to pretrain the model on the unlabeled data \cite{jing2020self}. These pretext tasks were hand-crafted based on domain knowledge, which includes rotation \cite{gidaris2018unsupervised}, solving jigsaw puzzles \cite{noroozi2016unsupervised}, relative patch prediction \cite{doersch2015unsupervised}, and so on. Many of these tasks are based on ad-hoc heuristics that limit the applicability of these approaches to broader domains. Consequently, a new family of SSL methods based on contrastive learning \cite{chen2020simple, he2020momentum} has emerged as the top-performing method that demonstrated excellent performance on many downstream tasks.  More recently, these techniques have been extended to medical image analysis \cite{azizi2021big, bai2019self, ciga2020self, koohbanani2021self, sowrirajan2020moco, srinidhi2021self} and have shown a promising viable alternative to fully supervised based methods. 

In the context of histopathology, a few domain-specific pretext tasks \cite{koohbanani2021self, srinidhi2021self} have been proposed to leverage multi-resolution contextual features for learning representations in pathology images. Notably, the recently proposed resolution sequence prediction (RSP) \cite{srinidhi2021self} pretext task has shown promising results on three different histopathology tasks, including patch-wise and slide-level classification problems. Contrastive learning based methods such as SimCLR have also been extended to histology \cite{ciga2020self} and have shown SOTA performance on many diverse histology tasks. However, in recent studies \cite{srinidhi2021self, yan2020clusterfit, zhai2019s4l}, it is shown that the representations learned by SSL methods are often overfitted to the pretraining objective and do not generalize well to downstream tasks. Furthermore, the improved efficiency of these methods is heavily dependent on the quantity of both labeled and unlabeled data \cite{cole2021does, reed2021self}, and most importantly, the aggressive data augmentation strategies \cite{li2020self, purushwalkam2020demystifying, xiao2020should} that are used during pretraining. Consequently, some recent works have attempted to improve the pretrained representations by leveraging hard examples either during the pretraining stage \cite{li2020self, robinson2020contrastive} or during the fine-tuning stage \cite{andreassen2021evolution, chen2020adversarial}. Inspired by these previous works, we propose to improve SSL on downstream tasks with a curriculum based hard example fine-tuning. We will show that our proposed technique improves robustness on both in-domain and out-of-domain distribution data, and furthermore, our method is generic and easily adaptable to any self-supervised pretrained objective.  

\vspace{1.5mm}
\noindent
\textbf{Curriculum Learning.}
The human learning mechanism follows a curriculum to understand complex tasks by imposing the order at which the complexity of the data is presented to the learner. For instance, human teachers often divide complex tasks into smaller sub-tasks and teach easier concepts first, followed by difficult concepts to another human. However, in machine learning, the supervision is often random, and training models have no clue about the difficulty of the sample which is being presented. One of the early seminal works by Bengio et al. \cite{bengio2009curriculum} demonstrated the applicability of CL to machine learning and showed that the learning improves if the data is presented in a meaningful order, with a gradual increase in complexity (typically, easier to hard). Following this intuition, several methods \cite{hacohen2019power, weinshall2018curriculum, wu2020curricula, zhou2020curriculum} have been proposed to determine the difficulty of the data sample and also the order in which it is presented to the network. Most of these previous methods either depend on the confidence of a pretrained model \cite{weinshall2018curriculum, hacohen2019power} or human-annotators to determine sample difficulty \cite{wei2021learn}. For instance, Wei et al. \cite{wei2021learn} explored the CL in histology based on multiple annotator agreements as a proxy to estimate the difficulty of a training sample. Notably, Wu et al. \cite{wu2020curricula} investigated several benefits of CL and provided thorough insights on when and where curriculum works to improve machine learning models on standard benchmark datasets. Our work takes inspiration from \cite{weinshall2018curriculum} and extends the idea of the curriculum by transfer learning to improve SSL on downstream tasks by generalizing representations to both in-domain and out-of-domain distribution data. 

\vspace{-1mm}
\section{Method}
\label{sec:Method}
Our approach consists of the following steps. First, we perform self-supervised pretraining on unlabeled data to learn histology specific visual representations. Second, we fine-tune the pretrained representations using hard examples via hardness-aware dynamic curriculum learning (HaDCL) approach. The HaDCL comprises two following stages: i) we first fine-tune the model with easy-to-hard examples (i.e., Curriculum-I stage), and ii) we then initialize the Curriculum-I model to fine-tune with hard-to-very-hard samples in the Curriculum-II stage. The details are presented next.

\subsection{Self-supervised Pretraining}
\label{ssec:Self-supervised Pretraining}
The goal of SSL is to first learn general visual representations with task-agnostic pretraining using unlabeled data. The pretraining is performed via solving a pretext objective, where the labels needed to train a convolutional neural network are generated within the data itself. These pretrained representations are transferred to downstream tasks by supervised fine-tuning on limited label data. In this work, we consider two prominent SSL techniques: a context-based Resolution Sequence Prediction (RSP) \cite{srinidhi2021self} and a contrastive learning based Momentum Contrastive Coding (MoCo) \cite{he2020momentum} approach. Our motivation behind adopting RSP and MoCo is because these techniques have shown consistent and reliable performance across a variety of histopathology tasks based on a recent study in \cite{srinidhi2021self}. 

\subsection{Hardness-aware Dynamic Curriculum Learning (HaDCL)}
\label{ssec:Hardness-aware Dynamic Curriculum Learning (HaDCL)}
In this section, we begin by answering the two following questions in the context of CL: \textbf{Q1.} \textit{How to measure the hardness or difficulty of a training sample?} ~and ~\textbf{Q2.} \textit{How to specify the order at which the training data is presented to the network?}. Before we begin, we shall setup some basic notations and definitions of CL. 

Let $D = \{(x_{i}, y_{i})\}_{i=1}^N$ denote the training data, where $x_{i} \in \mathbb{R}^{d}$ denotes an input sample and $y_{i} \in \mathcal{C}$ its corresponding label. In CL, the common approach is to train a target model $f_{\theta} : \mathcal{X} \rightarrow \mathcal{Y}$ with a set of non-uniformly sampled mini-batches $[\mathbb{B}_{1},...,\mathbb{B}_{M}] \subseteq D$ using a Stochastic Gradient Descent (SGD) optimization. To measure the instantaneous hardness of a training sample (\textbf{Q1}), we define a scoring function via curriculum by transfer learning approach \cite{hacohen2019power}: $s(x_{i}, y_{i}) \in \mathbb{R}$ based on the loss value $\ell=(f_{\theta}(x_{i}), y_{i})$ obtained from a pretrained self-supervised model ($f_{pre}(\cdot; \theta)$). We measure this instantaneous hardness of training samples during downstream fine-tuning by initializing $f_{\theta}$ with $f_{pre}(\cdot; \theta)$. We say that a sample $x_{j}$ is more difficult/hard than $x_{i}$, if $\ell(f_{\theta}(x_{j}), y_{j}) > \ell(f_{\theta}(x_{i}), y_{i})$. In this work, we consider $\ell(.,.)$ as the standard categorical cross-entropy loss to measure the instantaneous hardness of a training sample. 

Unlike in the previous work \cite{wei2021learn}, our proposed approach is more reliable to the training dynamics of a neural network; since it makes use of a powerful pretrained SSL model to examine the sample difficulty, which reflects the true underlying hardness of a training sample as it is experienced by the machine learner rather than a human-teacher. Such model-based ranking of training samples is of paramount importance in histopathology, where measuring hardness level by multiple annotator agreements is costly and sometimes infeasible for large-scale applications.

Next, we focus on the order in which the training data is presented to the network (\textbf{Q2}). Typically, an easy to hard (i.e., lowest to highest score ($s$)) strategy is followed to determine the ordering of samples during training. However, in the context of CL, we argue that there exist two main limitations: (i) due to randomness of SGD optimization, the instantaneous hardness of each training sample can vary significantly over consecutive epochs, which may not reflect the true hardness level of a sample over time with the model being trained. This is because the easier samples stay easy throughout training since their loss value is more likely to stay at samples minima; while for hard examples, the loss value is relatively less stable during the early part of the training and gradually stabilizes as we train more on them. Thus the instantaneous hardness level of a sample tends to decrease monotonically during training and cannot be at a fixed level; (ii) further, keeping track of the instantaneous hardness of each sample up-to-date requires extra inference computation over all training samples, which can be computationally challenging for neural networks \cite{jiang2019accelerating}.

The aforementioned limitations motivated us to propose a ``\textit{hardness-aware dynamic curriculum learning} (\textbf{HaDCL})" approach to dynamically determine the sample's instantaneous hardness level over the gradual course of training. Our proposed approach consists of a \textit{\textbf{dual-stage}} curriculum training strategy, which we apply during downstream fine-tuning. In the first stage, we focus on \textit{easy-to-hard} samples, and in the second stage, we focus on \textit{hard-to-very-hard} samples for fine-tuning the pretrained SSL model. 

In \textbf{Curriculum-I} (i.e., \textbf{\textit{easy-to-hard}}) stage, we first initialize the downstream fine-tuning model $f_{ft}(\cdot;\theta)$ with the pretrained SSL model $f_{pre}(\cdot;\theta)$, and compute loss for all input samples $\ell = (f_{ft}(x_{i}), y_{i})_{i=1}^B$ in a mini-batch $\mathbb{B}$ using categorical cross-entropy. Next, all $B$ samples within a mini-batch $\mathbb{B}$ are sorted in descending order by their loss value $\ell$ to obtain a set $\tilde{D}$. From the sorted set $\tilde{D}$, we select the top-$K$ samples that constitute the hard examples: \textbf{top-}$\boldsymbol{K}$ = $\boldsymbol{\alpha \times B}$, where $\alpha$ is parameter $(0 \leq \alpha \leq 1)$ which denotes the portion of hard samples in a set $\tilde{D}$. However, relying only on the portion of hard samples in each mini-batch does not always necessarily consider varying hardness levels between mini-batches. In other words, treating all mini-batches equally may lead to sub-optimal performance as different batches will have a varying number of hard examples. Thus, the level of hardness must be smoothly adjusted according to the training dynamics of neural network to account for varying instantaneous hardness levels of training samples over time. Therefore, we choose to \textit{dynamically} determine the mini-batch instantaneous hardness level using an adaptive \textbf{\textit{threshold}} as
\vspace{-3mm}
\begin{equation}
   \textit{thres} = a(1 - \frac{t}{T}) + b,
   \label{eq:threshold}
\end{equation}
where, $a$ and $b$ are hyperparameters (such that, $a \gg b$) which controls $\textit{thres}$ such that its value changes from $a+b \rightarrow b$ at uniform speed over the gradual course of training. The term $t$ denotes the current iteration, and $T$ indicates the total number of iterations within an epoch. 

Next, we dynamically update the model weights $\theta$ in a mini-batch $\mathbb{B}$ based on the top-$K$ samples in set $\tilde{D}$ for which the \textbf{sum of top-$\boldsymbol{K}$ loss} (i.e., $\sum_{k=1}^{K} \ell_{k}$) exceeds the threshold \textbf{\textit{`thres'}} as 
\vspace{-2.5mm}
\begin{equation}
  \sum_{k=1}^{K} \ell_{k} > \textit{thres} \times \sum_{i=1}^{B} \ell_{total},
  \label{eq:loss1}
      \vspace{-0.5mm}
\end{equation}
where, $\sum_{i=1}^{B} \ell_{total}$ is the \textbf{\textit{total loss}} value over all $B$ samples within a mini-batch $\mathbb{B}$. By doing so, we can avoid an extra inference step for keeping track of the instantaneous hardness of each sample up-to-date, which can be computationally expensive. Further, this dynamic way of updating the model parameters based on mini-batch hardness level can simultaneously alleviate both under-fitting (for hard samples) and over-fitting (for easy samples) problems. From Eq. \eqref{eq:threshold}, during the early phase of training (i.e., at $t \geq 1$), the model is oriented to learn with a larger number of easier examples - due to a larger threshold value \textit{thres} $\approx a+b$; while, during the later part of the training (i.e., at $t \approx T$) fewer but hard samples are learned - due to lower threshold value of \textit{thres} $\approx b$. This dynamic way of CL allows the model to revisit more frequently those samples that have been historically hard, while making less frequent revisits to those easier samples that have been already learned.

In \textbf{Curriculum-II} training, we start by initializing the model ($\theta_{2}$) with Curriculum-I fine-tuned model ($\theta_{1}$) and focus on \textbf{\textit{hard-to-very-hard}} examples for CL. Here, we determine the instantaneous hardness of hard to very-hard samples by dynamically choosing a subset of top-$K'$ samples, within a pre-defined set of top-$K$ samples as: \textbf{top-$\boldsymbol{K'}$ = $\textit{thres} ~\times$ top-$\boldsymbol{K}$}; where, \textit{thres} is an adaptive threshold (see, Eq. \eqref{eq:threshold}) to estimate the mini-batch instantaneous hardness level over top-$K'$ samples. 

In this stage, we dynamically update the fine-tuned Curriculum-I model weights $\theta_{1}$ based on top-$K'$ samples in a set $\tilde{D}$, for which the \textbf{sum of top-$\boldsymbol{K'}$ loss} (i.e., $\sum_{k'=1}^{K'} \ell_{k'}$) exceeds the threshold \textbf{\textit{`thres'}} as
\vspace{-1.5mm}
\begin{equation}
  \sum_{k'=1}^{K'} \ell_{k'} > \textit{thres} \times \sum_{k=1}^{K} \ell_{k},
  \label{eq:loss2}
      \vspace{-0.5mm}
\end{equation}
where, $\sum_{k=1}^{K} \ell_{k}$ is the sum of loss values over top-$K$ samples within a mini-batch $\mathbb{B}$, as defined in Eq. \eqref{eq:loss1}. The pseudocode for our proposed dual-stage HaDCL strategy is illustrated in Algorithm \ref{Algo:HaDCL method}.
\vspace{-1mm}

\begin{algorithm}
\footnotesize
\DontPrintSemicolon
\vspace{1mm}
\SetKwInOut{Parameter}{Inputs}
\Parameter{$D = \{(x_{i}, y_{i})\}_{i=1}^B$ = training samples in mini-batch $\mathbb{B}$\\
$f_{pre}(\cdot ;\theta)$ = SSL pretrained model \\
$\ell=(f_{\theta}(x_{i}), y_{i}) \in \mathbb{R}$ = loss / scoring function \\ 
$\mathit{o} \in$ \{``descending"\}  = order \\
$\alpha$ = portion of hard samples in a set $\tilde{D}$ \\
$a, b$ = hyperparameters such that $a \gg b$ \\
$t$ = current iteration \\
$T$ = total number of iterations within an epoch \\
$f_{ft}(\cdot ;\theta)$ = fine-tuning model \\
\vspace{1mm}
\Indm $\tilde{D} = (x_{1},..., x_{B}) \leftarrow$ \textbf{sort}$(\{x_{1},..., x_{B}\}, \ell, o)$
}

\vspace{2mm}
\SetKwInOut{Parameter}{\textcolor{blue}{Curriculum-I stage}}
\Parameter{}
\vspace{0.5mm}
\textit{\textbf{Initialize:}} $f_{ft}(\cdot ;\theta_1) \leftarrow f_{pre}(\cdot ;\theta)$, with weights $\theta$ unfrozen across entire network + a 2-layer MLP (Fc1, ReLU, Fc2) that predicts class logits \\
\vspace{0.5mm}
\For{$epoch$ in $[1,...,num\_epochs]$\vspace{0.25mm}}{
\For{each minibatch $\mathbb{B}_{t}$, where $t \in [1, ..., T]$\vspace{0.25mm}}{
top-$K$ = $\alpha \times B$\\
$thres = a(1 - \frac{t}{T}) + b$ ;~~adaptive hardness threshold  \\
$D' = \tilde{D}[0:$ top-$K]$ ;~~top-$K$ \textit{\textbf{hard}} samples\\
$\ell_{total} = \ell(D) = \ell(f_{ft}(x_{i}), y_{i})_{i=1}^B$ ;~\textbf{total loss}\\
$\ell_{k} = \ell(D') = \ell(f_{ft}(x_{k}), y_{k})_{k=1}^K$ ;~~\textbf{top-$K$ loss}\\
\vspace{1mm}
\eIf{$\sum_{k=1}^{K} \ell_{k} > thres \times \sum_{i=1}^{B} \ell_{total}$ \vspace{0.5mm}}
{Update ~$f_{ft}(\cdot ;\theta_1)$ with $\ell_{k}$}
{Update ~$f_{ft}(\cdot ;\theta_1)$ with $\ell_{total}$}
}}
\textbf{return} $\theta_1$;

\vspace{2mm}
\SetKwInOut{Parameter}{\textcolor{blue}{Curriculum-II stage}}
\Parameter{}
\vspace{0.5mm}
\textit{\textbf{Initialize:}} $f_{ft}(\cdot ;\theta_2) \leftarrow \theta_1$ \\
\vspace{0.5mm}
\For{$epoch$ in $[1,...,num\_epochs]$\vspace{0.25mm}}{
\For{each minibatch $\mathbb{B}_{t}$, where $t \in [1, ..., T]$\vspace{0.25mm}}{
$thres = a(1 - \frac{t}{T}) + b$ ;~~adaptive hardness threshold\\
top-$K$ = $\alpha \times B$\\
top-$K'$ = $thres ~\times$ top-$K$\\
$D' = \tilde{D}[0:$ top-$K]$ ;~~top-$K$ \textit{\textbf{hard}} samples\\
$D'' = D'[0:$ top-$K']$ ;~~top-$K'$ \textit{\textbf{very-hard}} samples\\
$\ell_{k} = \ell(D') = \ell(f_{ft}(x_{k}), y_{k})_{k=1}^K$;~\textbf{top-$K$ loss}\\
$\ell_{k'} = \ell(D'') = \ell(f_{ft}(x_{k'}), y_{k'})_{k'=1}^{K'}$; \textbf{top-$K'$ loss}\\
\vspace{1mm}
\eIf{$\sum_{k'=1}^{K'} \ell_{k'} > thres \times \sum_{k=1}^{K} \ell_{k}$\vspace{0.5mm}}
{Update ~$f_{ft}(\cdot ;\theta_{2})$ with $\ell_{k'}$}
{Update ~$f_{ft}(\cdot ;\theta_{2})$ with $\ell_{k}$}
}}
\textbf{return} $\theta_2$;
\caption{HaDCL method}
\label{Algo:HaDCL method}
\end{algorithm}

\section{Experiments}
\label{sec:Experiments}
In this section, we validate our method on \textbf{three} standard benchmark datasets for breast cancer metastasis detection in lymph nodes at whole-slide-image (WSI)-level (\textbf{Camelyon16, MSK}) \cite{bejnordi2017diagnostic, campanella2019clinical} and patch-level colorectal polyps classification (\textbf{MHIST}) \cite{wei2021learn}. We choose these three datasets to investigate the relative benefits of CL on standard high and low-data training regimes. In addition, the chosen tasks embody both patch-wise and slide-level classification in histopathology and explore the problem of domain shift when training data from the target domain is entirely absent. This helps to understand the generalizability of our proposed approach and the boundaries within which the CL works to improve SSL in practice.    

\subsection{Datasets}
\label{ssec:Datasets}
We first perform self-supervised pretraining on the Camelyon16 dataset, followed by fine-tuning the pretrained model with our proposed HaDCL approach on Camelyon16 and MHIST datasets, respectively; and finally, evaluated on test sets of three datasets: Camelyon16, MSK, and MHIST. We will next introduce the datasets in detail.

\vspace{1mm}
\textbf{Camelyon16 dataset \cite{bejnordi2017diagnostic}.} Camelyon16 consists of 399 hematoxylin and eosin (H\&E) stained WSIs (from 399 patients) of lymph nodes in the breast, divided into 270 for training and 129 for testing. The WSIs were acquired from two different centers using two different scanners with specimen level pixel sizes of ($0.226 \mu m/pixel$) and ($0.243 \mu m/pixel$). For self-supervised pretraining, we only considered 60 WSIs (slide id: normal set (1-35); tumor set (1-25)) from the total 270 training images discarding their labels (we refer to this as an \textbf{\textit{unlabeled} set}). While, the downstream fine-tuning is performed with 228 WSIs (85\%) (slide id: normal set (1-135); tumor set (1-93)) and validation with the rest 42 WSIs (15\%) (slide id: normal set (136-160); tumor set (94-110)). Further, the fine-tuning set contains 400K patches (200K tumor and 200K normal), and the validation set contains 40K patches (20K tumor and 20K normal). The test set contains an independent set of 129 WSIs (49 with nodal metastases and 80 normal WSIs). 

\vspace{1mm}
\textbf{MSK dataset \cite{Campanella_Hanna_Brogi_Fuchs_2019}.} MSK set was released as part of a previous study in \cite{campanella2019clinical}, which contains an independent test set of 130 H\&E stained WSIs of axillary lymph nodes from 78 breast cancer patients. The nodal metastasis is present in 36 images from 27 patients with corresponding slide-level labels. The WSIs were scanned at $20\times$ magnification ($0.5 \mu m/pixel$). Note: the publicly released dataset\footnote{\textcolor{red}{https://doi.org/10.7937/tcia.2019.3xbn2jcc}} is only an independent test set and does not contain training images. MSK is considered out-of-distribution (OOD) to Camelyon because of three reasons: i) image resolution difference ($20 \times$ in MSK vs. $40 \times$ magnification in Camelyon); ii) technical variability in slide preparation \cite{campanella2019clinical}; iii) presence of cases with signs showing the effect of treatment response from neoadjuvant chemotherapy in MSK vs. no treatment response cases in Camelyon. Therefore, we choose the MSK dataset to test the generalizability of our approach to domain shift. 

\vspace{1mm}
\textbf{MHIST dataset \cite{wei2021learn}.} MHIST contains a total of 3,152 images (with $224 \times 224$ pixels) for classifying colorectal polyps as between hyperplastic polyps (HPs) and sessile serrated adenomas (SSAs). This dataset is split into a training set consisting of 2,175 images, whereas the test set contains 977 images. We further divide the train set into fine-tuning set with 1740 images (80\%) and a validation set of 435 images (20\%). Multiple annotators annotated the images, and majority voting of labels was performed to obtain the final ground truth.

\subsection{Implementation Details}
\label{ssec:Implementation Details}
We first perform \textbf{self-supervised pretraining} of RSP and MoCo on Camelyon16 \textit{unlabeled} set using ResNet-18 as our base encoder network. We adopt similar hyperparameter settings and domain-specific data augmentation strategies for RSP and MoCo pretraining as reported in \cite{srinidhi2021self}. After pretraining, we only use the ResNet encoder $f_{\theta}$ (that maps output to a 512-dimensional embedding) for downstream fine-tuning, discarding the project head (2-layer MLP in previous design \cite{srinidhi2021self}) following the suggestion in SimCLR \cite{chen2020simple}. Further, we choose to fine-tune from the first layer in the encoder $f_{\theta}$ with a newly initialized 2-layer MLP (Fc1, ReLU, Fc2) that predicts the class logits for final classification using all labeled samples and a standard supervised cross-entropy loss. 

In our experiments, we fine-tune the pretrained model with a patch size of $256 \times 256$ pixels on Camelyon16 and MHIST datasets ($224 \times 224$ resized to $256 \times 256$), respectively, followed by evaluation on Camelyon16, MSK, and MHIST test sets. Note: to account for the input resolution differences between MSK ($0.5 \mu m/pixel$) and Camelyon16 datasets ($0.23 - 0.24 \mu m/pixel$), we choose to test the camelyon16 fine-tuned model on MSK by upsampling the input patch from $256 \times 256$ to $512 \times 512$ pixels, followed by centre cropping to $256 \times 256$ pixels. For fine-tuning, we use the following sets of domain-specific data augmentations \cite{tellez2019quantifying}: perturbations of hue and saturation values between (-0.1, 0.1) and (-1, 1), respectively in HSV color space, additive Gaussian noise with $\mu=0$ and $\sigma=(0, 0.1)$, shifting brightness and contrast intensity ratios between (-0.2, 0.2), blurring with a random-sized kernel (3, 7), affine transformation with translation, scale and rotation limit of ($0.0625, 0.5, 45^{\circ}$), rotation with centre crop of (-$90^{\circ}, +90^{\circ}$) and finally, we scale with a factor of $(0.8, 1.2)$ and randomly resize and crop the image patch to its original size. We apply these augmentations in sequence by randomly selecting 2 of total 7 augmentations in each mini-batch, similar to RandAugment technique \cite{cubuk2020randaugment}.

The \textbf{fine-tuning} is performed with three different strategies: \textbf{supervised fine-tuning} (\textit{vanilla baseline}), \textbf{curriculum-I} and \textbf{curriculum-II} fine-tuning as described in Section. \ref{ssec:Hardness-aware Dynamic Curriculum Learning (HaDCL)}. We first list the hyperparameters common to all three strategies for \textbf{Camelyon16} dataset: we set the batch size to 512 and optimize the network with Adam optimizer ($\beta_{1} = 0.9$, $\beta_{2} = 0.999$) with a weight decay of $1e-4$. Next, we train the model for 250 epochs, with an initial learning rate ($lr$) of $5e^{-4}$ and a multi-step decay at (60, 120, 180) epochs by 0.1 for supervised and Curriculum-I fine-tuning; while we train for 60 epochs with $lr$=$1e^{-5}$ and a multi-step decay at 30$^{th}$ epoch by 0.1 for Curriculum-II stage. We set empirically the parameters $k$ as 0.10 and $a$ and $b$ in Eq. \ref{eq:threshold} as (0.7, 0.2) in both Curriculum-I and II stages (refer, Section \ref{sssec:Ablation Study} for ablations). For the \textbf{MHIST} dataset, we adopted the same settings as Camelyon16, except the following parameters: we set the batch size as 32 and trained for 1000 epochs with $lr$=$1e^{-5}$ and a multi-step decay at (200, 400, 600, 800) epochs by 0.95 for supervised and Curriculum-I fine-tuning; while for Curriculum-II, we trained for 60 epochs with $lr$=$1e^{-5}$ and a multi-step decay at 30$^{th}$ epoch by 0.95. Finally, we saved the best model based on the highest validation accuracy to test on the test set. We implemented our approach in PyTorch and trained with Nvidia V100 GPUs.

\begin{table*}
\Large
\centering
\caption{WSI classification results on Camelyon16 and MSK set evaluated with WSI-level accuracy and area under the curve (AUC) with 95\% CIs shown in square brackets; followed by patch-wise colorectal polyps classification on MHIST dataset evaluated with patch-level accuracy and AUC. The DeLong method \cite{sun2014fast} was used to construct 95\% CIs. The best results for each dataset are highlighted in bold.}
\label{tab:overall_results}
\vspace{1.5mm}
\resizebox{0.75\linewidth}{!}{
\begin{tabular}{@{}lllcccccc@{}}
\toprule \toprule[1pt]
\multicolumn{2}{c}{\multirow{2}{*}{\textbf{Pretraining}}}                        &
\multicolumn{1}{l}{\multirow{2}{*}{\textbf{Fine-tuning}}}                        &
\multicolumn{2}{c}{\textbf{Camelyon16} (slide-level)} & \multicolumn{2}{c}{\textbf{MSK} (slide-level)} & \multicolumn{2}{c}{\textbf{MHIST} (patch-level)} \\ \cmidrule(l){4-9} 
\multicolumn{3}{c}{}                                      & \textbf{Accuracy} & \textbf{AUC} & \textbf{Accuracy} & \textbf{AUC} & \textbf{Accuracy} & \textbf{AUC} \\ \midrule
\multicolumn{2}{l}{\multirow{3}{*}{\textbf{Random}}} & Baseline           & 0.760              & 0.780 [0.595-0.804] & 0.285          & 0.518 [0.403-0.632] & 0.803           & 0.880           \\ \cmidrule(l){3-9} 
\multicolumn{2}{l}{}                      & \textbf{Curriculum-I}  & 0.853  & 0.814 [0.735-0.892]   & 0.400        & 0.685 [0.571-0.798] & 0.802        & 0.889   \\ \cmidrule(l){3-9} 
\multicolumn{2}{l}{}                      & \textbf{Curriculum-II} & 0.822        & 0.845 [0.768-0.922]   & 0.800        & 0.744 [0.645-0.842] & 0.795  & 0.874   \\ \midrule [1pt]
\multicolumn{2}{l}{\multirow{3}{*}{\textbf{RSP \cite{srinidhi2021self}}}}  & Baseline      & 0.752        &  0.806 [0.724-0.887] &    0.285      &   0.542 [0.428-0.654]  &    0.816      & 0.888   \\ \cmidrule(l){3-9} 
\multicolumn{2}{l}{}                      & \textbf{Curriculum-I}  & 0.860        &  0.891 [0.824-0.958]   &    0.654      &  0.743 [0.650-0.835]   &    0.805      & 0.880   \\ \cmidrule(l){3-9} 
\multicolumn{2}{l}{}                      & \textbf{Curriculum-II} & \textbf{0.891}        & \textbf{0.942 [0.897-0.987]} &   0.669      &   0.771 [0.670-0.871]  &    0.793      & 0.872   \\ \midrule [1pt]
\multicolumn{2}{l}{\multirow{3}{*}{\textbf{MoCo \cite{he2020momentum}}}} & Baseline      &   0.744      &  0.837 [0.759-0.915]  &  \textbf{0.846}        &  0.749 [0.645-0.852]   &    0.815      & 0.884    \\ \cmidrule(l){3-9} 
\multicolumn{2}{l}{}                      & \textbf{Curriculum-I}  &   0.729      &  0.829 [0.751-0.906]  &   0.823       &   \textbf{0.771 [0.667-0.874]}  &   \textbf{0.825}       &  \textbf{0.896}  \\ \cmidrule(l){3-9} 
\multicolumn{2}{l}{}                      & \textbf{Curriculum-II} &  0.744      & 0.854 [0.783-0.925]   &     0.808    &  0.771 [0.676-0.864]  &   0.815      & 0.887  \\ \bottomrule \bottomrule[1pt]
\end{tabular}}
\vspace{-12mm}
\end{table*}

\subsection{Results and Discussion}
\label{ssec:Results and Discussion}
We validate the performance of our HaDCL approach with strong set of baselines: (i) \textbf{pretraining} with \textbf{RSP} \cite{srinidhi2021self} and \textbf{MoCo} \cite{he2020momentum} based SSL methods, along with \textbf{fully-supervised} method (\textbf{\textit{randomly}} initialized); (ii) \textbf{fine-tuning} with 3 different strategies: \textbf{Supervised (\textit{`Baseline'}}, as depicted in Table \ref{tab:overall_results}), \textbf{Curriculum-I} (with \textit{easy-to-hard} examples) and \textbf{Curriculum-II} (with \textit{hard-to-very-hard} examples). We evaluate these baselines for breast cancer metastasis detection at WSI-level (Camelyon16, MSK) and patch-level colorectal polyps classification (MHIST) tasks. For WSI-level classification, a random-forest-based slide-level classifier was used to obtain the final slide-level predictions. Similar to Wang et al. \cite{wang2016deep}, we extract several geometrical features from the heatmap predictions (connected component analysis with threshold of 0.5 and 0.95) to train a final slide-level classifier. We used accuracy (Acc) and area under the receiver operating characteristic curve (AUC) as evaluation metrics for accessing both WSI-level and patch-level classification performance. Further, to check whether our HaDCL approach significantly improved the performance, we also computed statistical significance test using Delong's test \cite{sun2014fast} for pairs of AUCs between supervised (baseline) and HaDCL based fine-tuning methods. The 95\% CIs were computed to access the significance at $p$-value $< 0.05$.

\begin{figure*}
\centering
\begin{tabular}{
  @{}>{\centering\arraybackslash}m{\dimexpr0.92\textwidth-\tabcolsep\relax}
  >{\centering\arraybackslash}m{\dimexpr.0001\textwidth-\tabcolsep\relax}@{}
  }
 	\begin{minipage}{0.245\linewidth}
 		\centerline{\includegraphics[height = 1.6cm, width = 3.9cm]{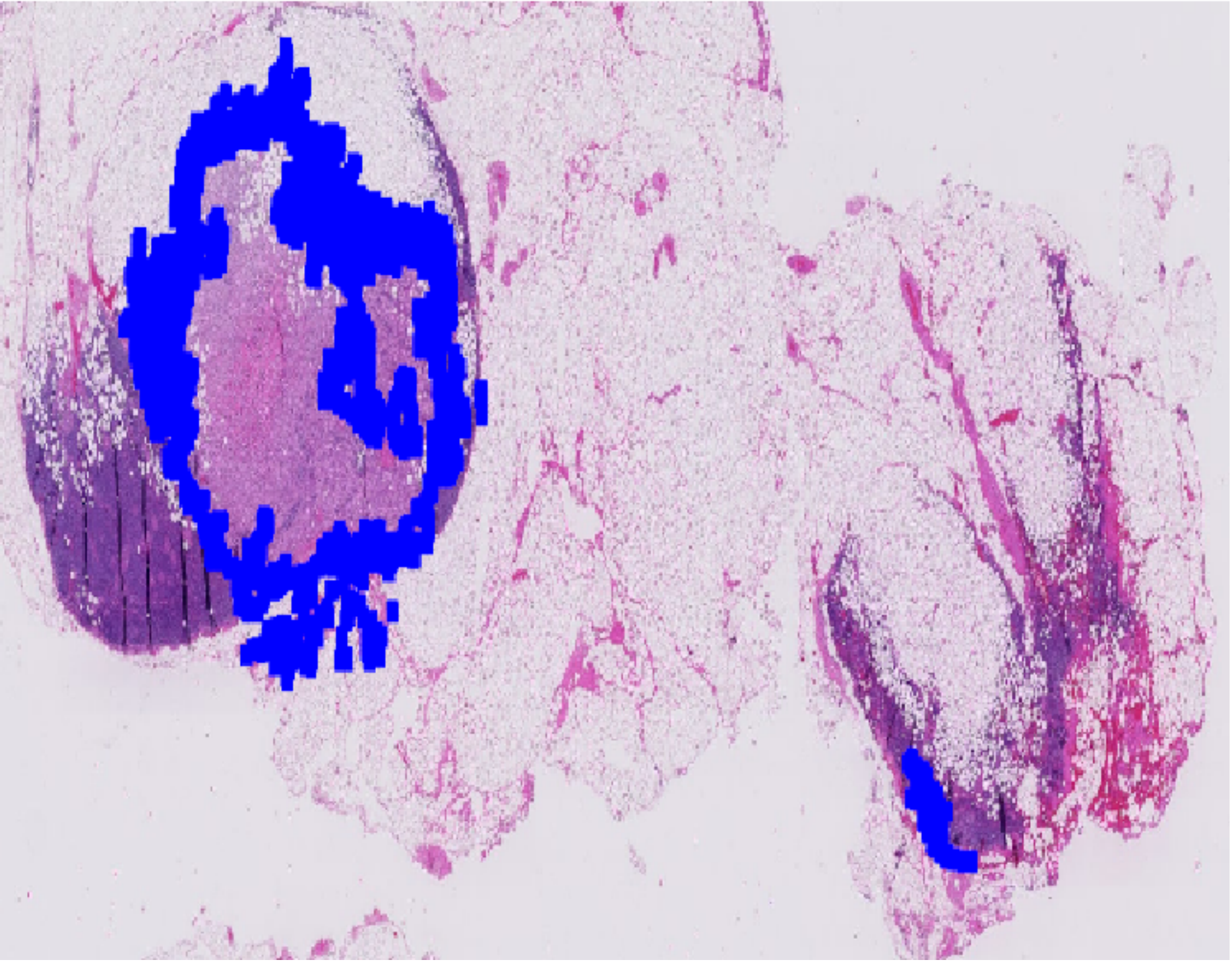}}
	\end{minipage}
    \hfill
    \begin{minipage}{0.245\linewidth}
 		\centerline{\includegraphics[height = 1.6cm, width = 3.9cm]{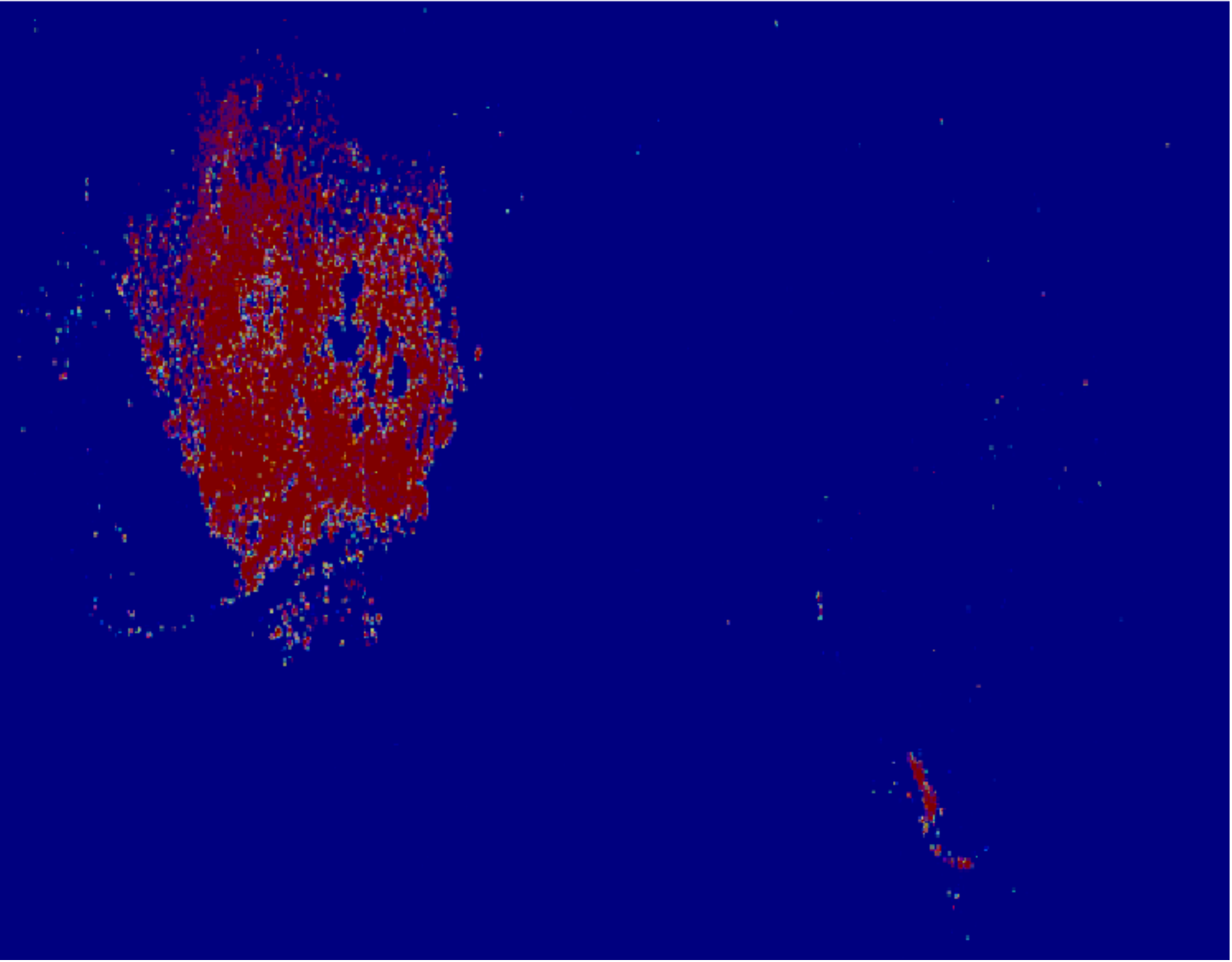}}
 	\end{minipage}
    \hfill
    \begin{minipage}{0.245\linewidth}
 		\centerline{\includegraphics[height = 1.6cm, width = 3.9cm]{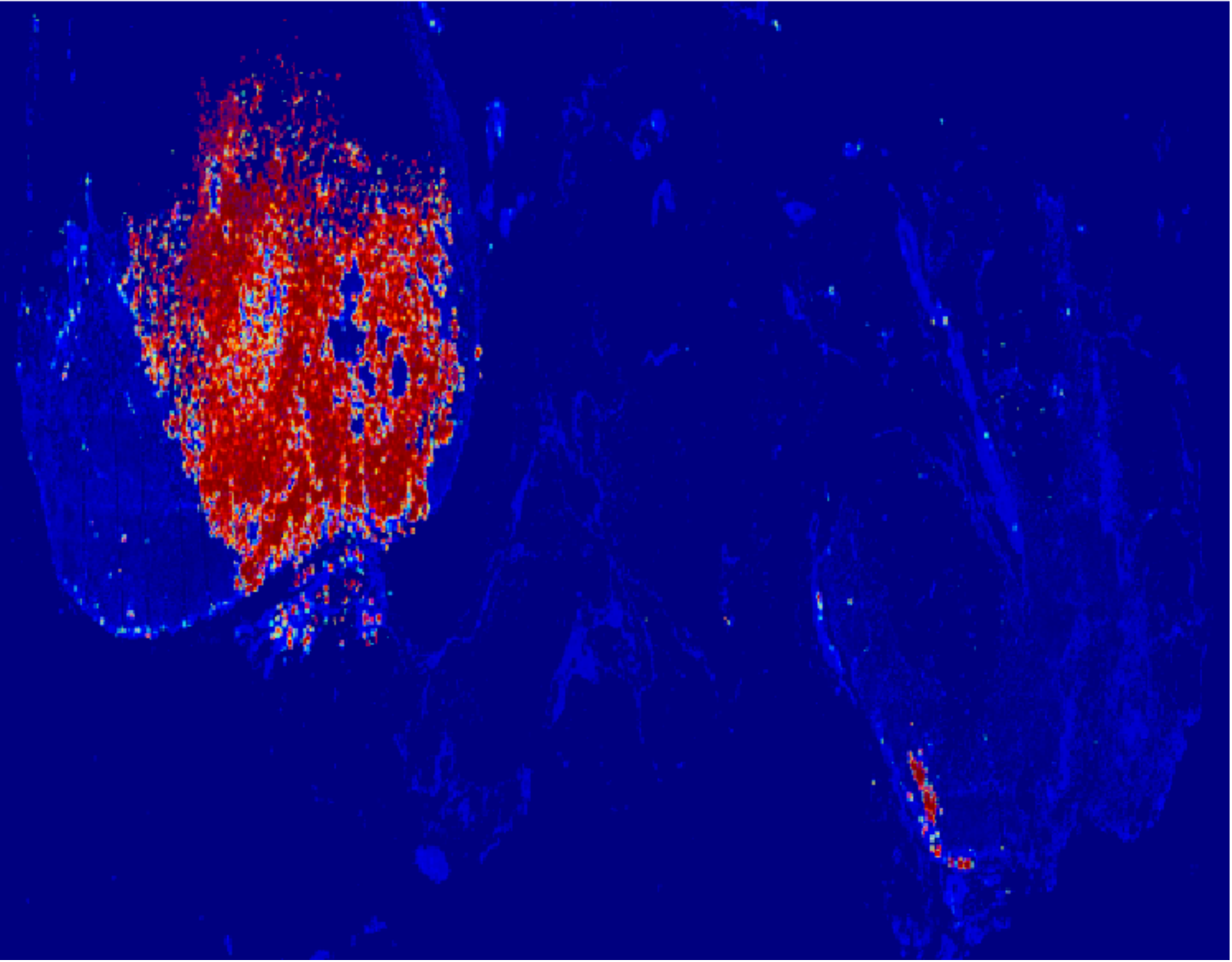}}
 	\end{minipage}
    \hfill
    \begin{minipage}{0.245\linewidth}
 		\centerline{\includegraphics[height = 1.6cm, width = 3.9cm]{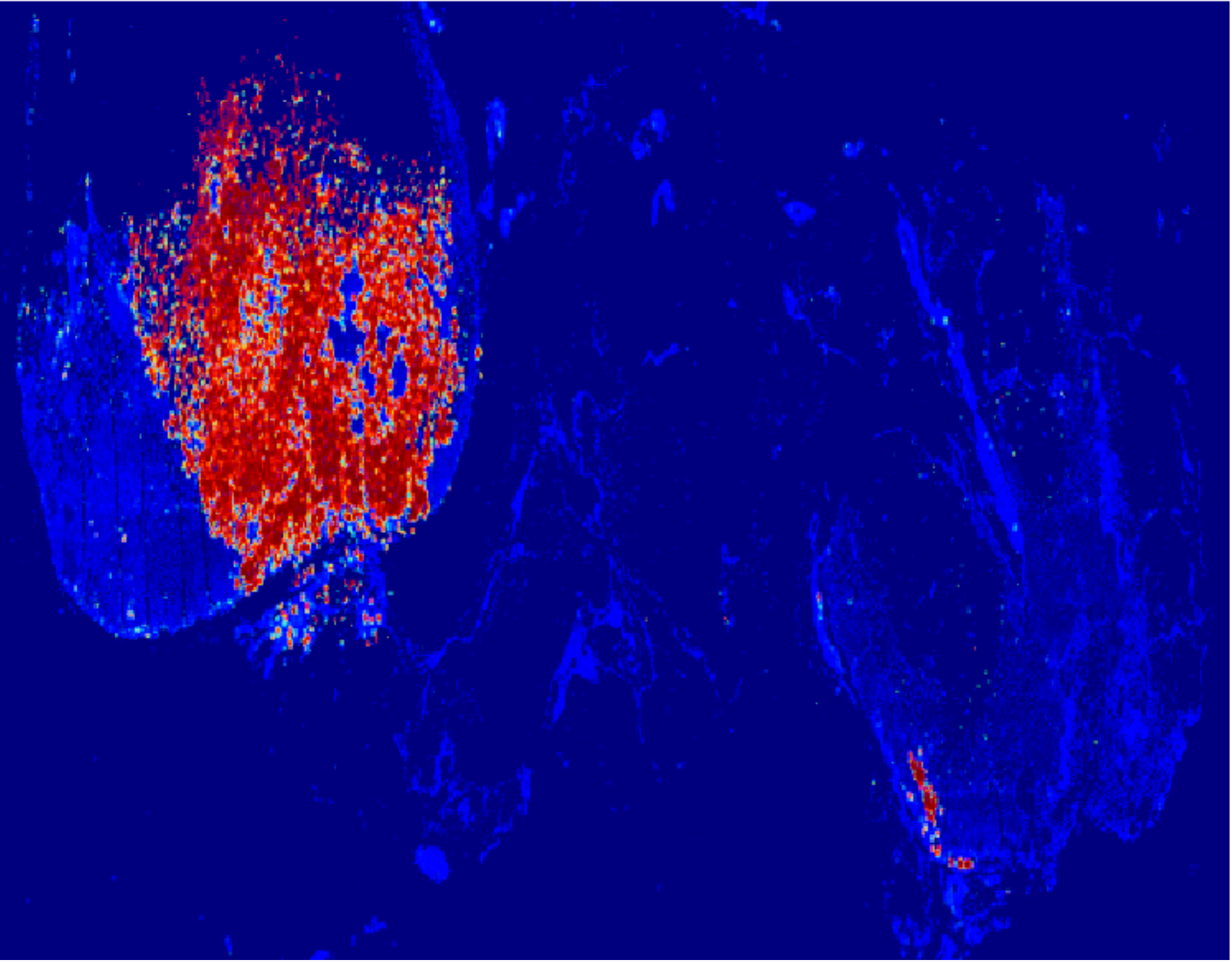}}
 	\end{minipage}
 	\vfill
    \begin{minipage}{0.245\linewidth}
 		\centerline{\includegraphics[height = 1.6cm, width = 3.9cm]{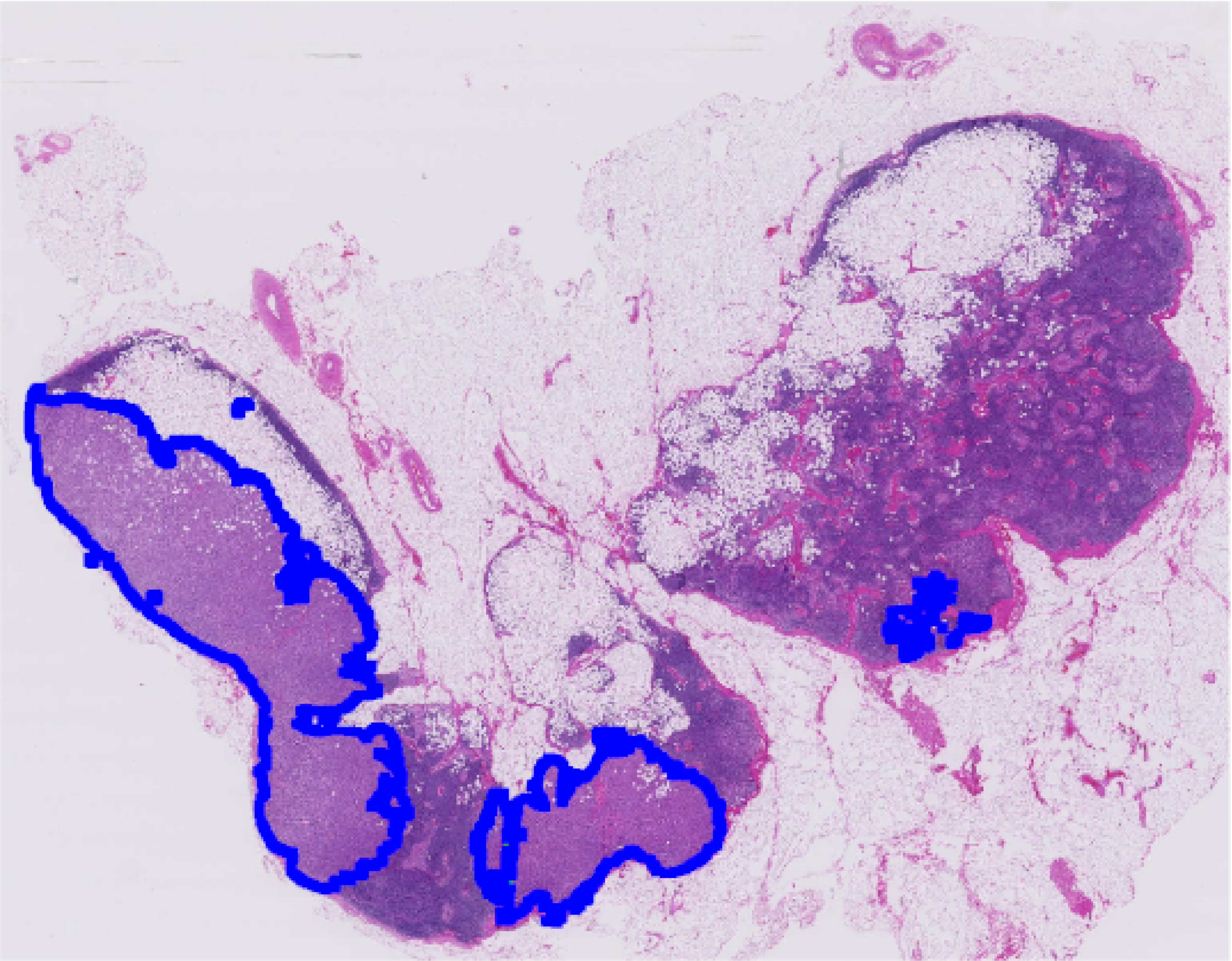}}
 		\centerline{(a) Original}
	\end{minipage}
    \hfill
    \begin{minipage}{0.245\linewidth}
 		\centerline{\includegraphics[height = 1.6cm, width = 3.9cm]{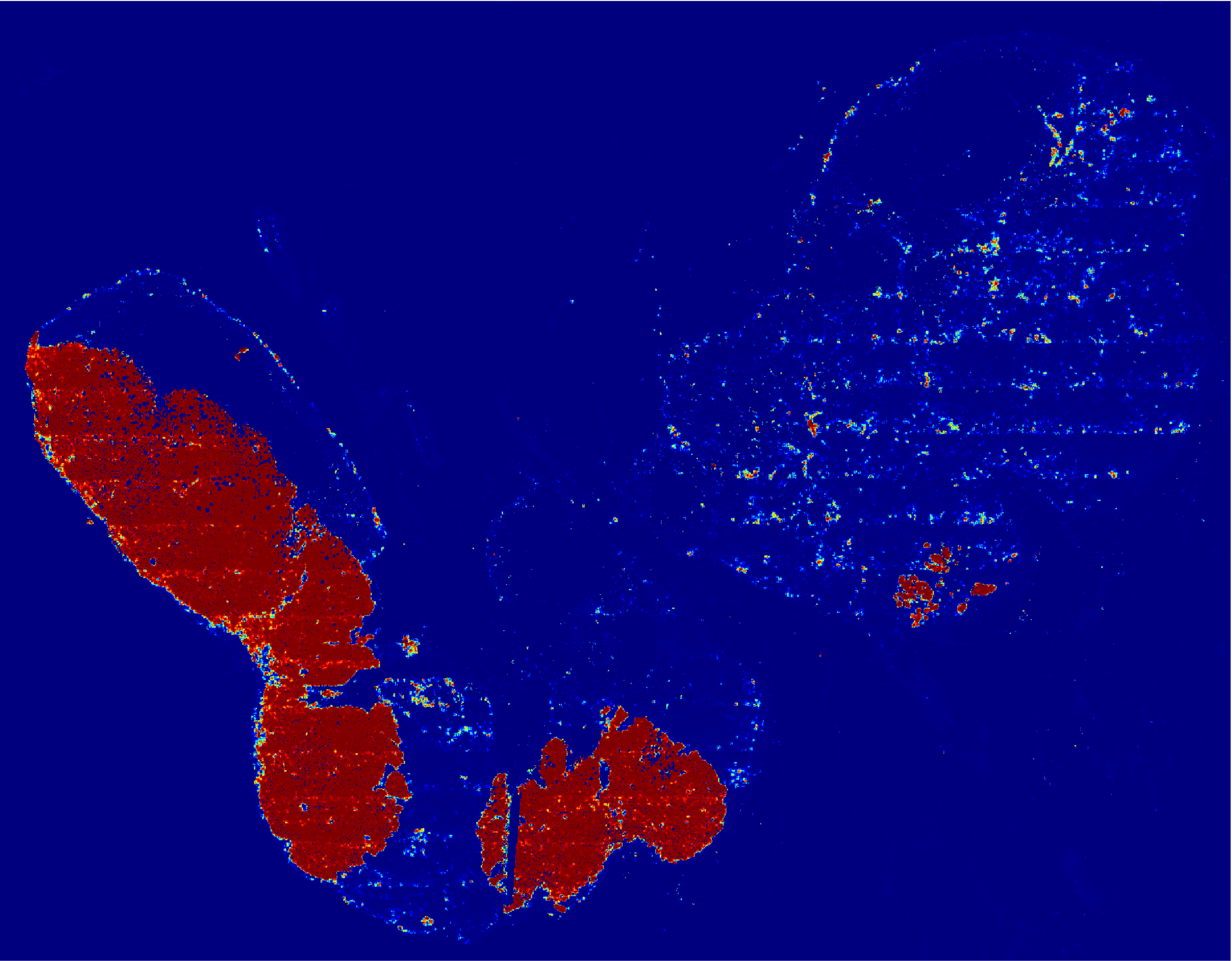}}
 		\centerline{(b) Baseline}
 	\end{minipage}
    \hfill
    \begin{minipage}{0.245\linewidth}
 		\centerline{\includegraphics[height = 1.6cm, width = 3.9cm]{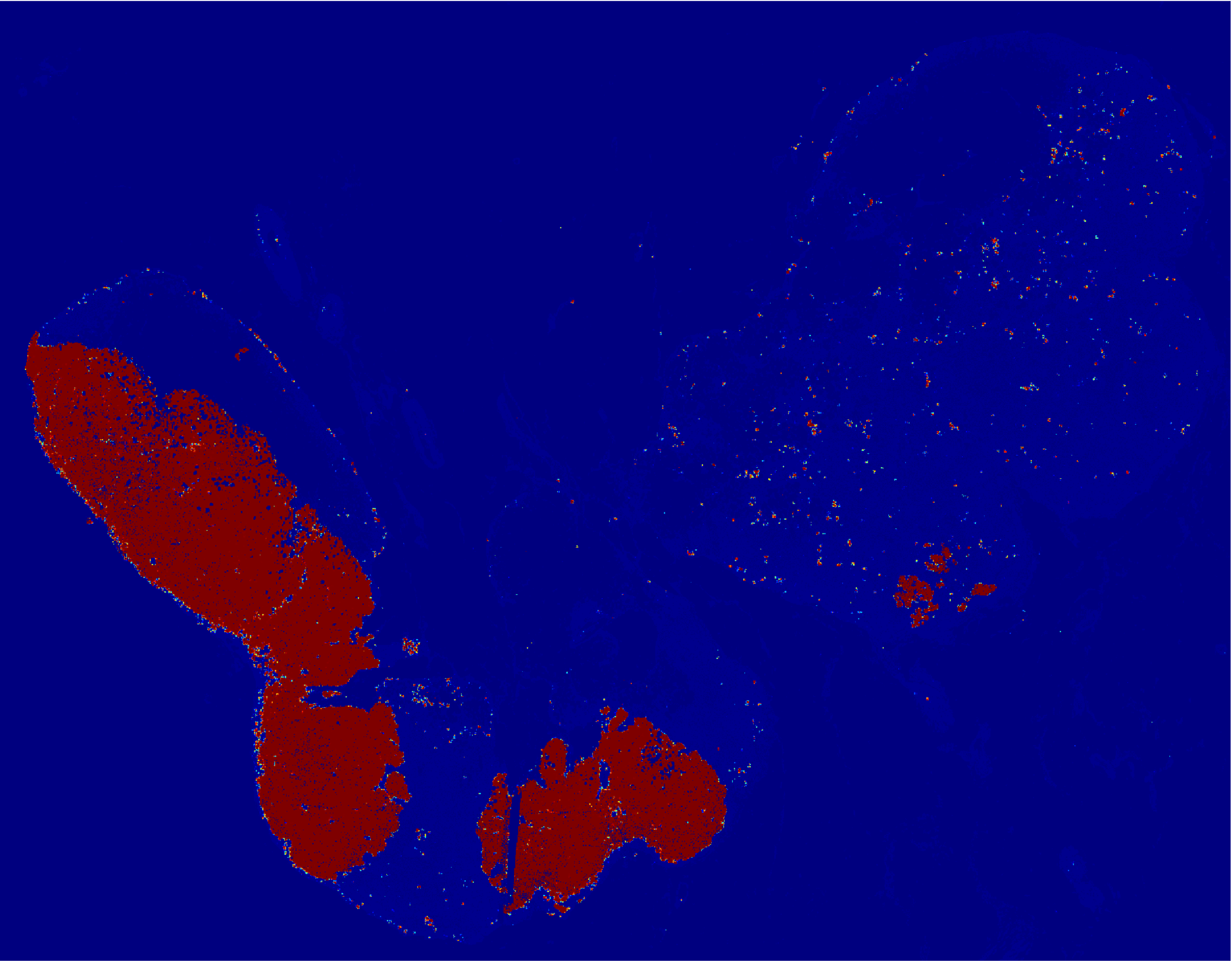}}
 		\centerline{(c) Curriculum-I}
 	\end{minipage}
    \hfill
    \begin{minipage}{0.245\linewidth}
 		\centerline{\includegraphics[height = 1.6cm, width = 3.9cm]{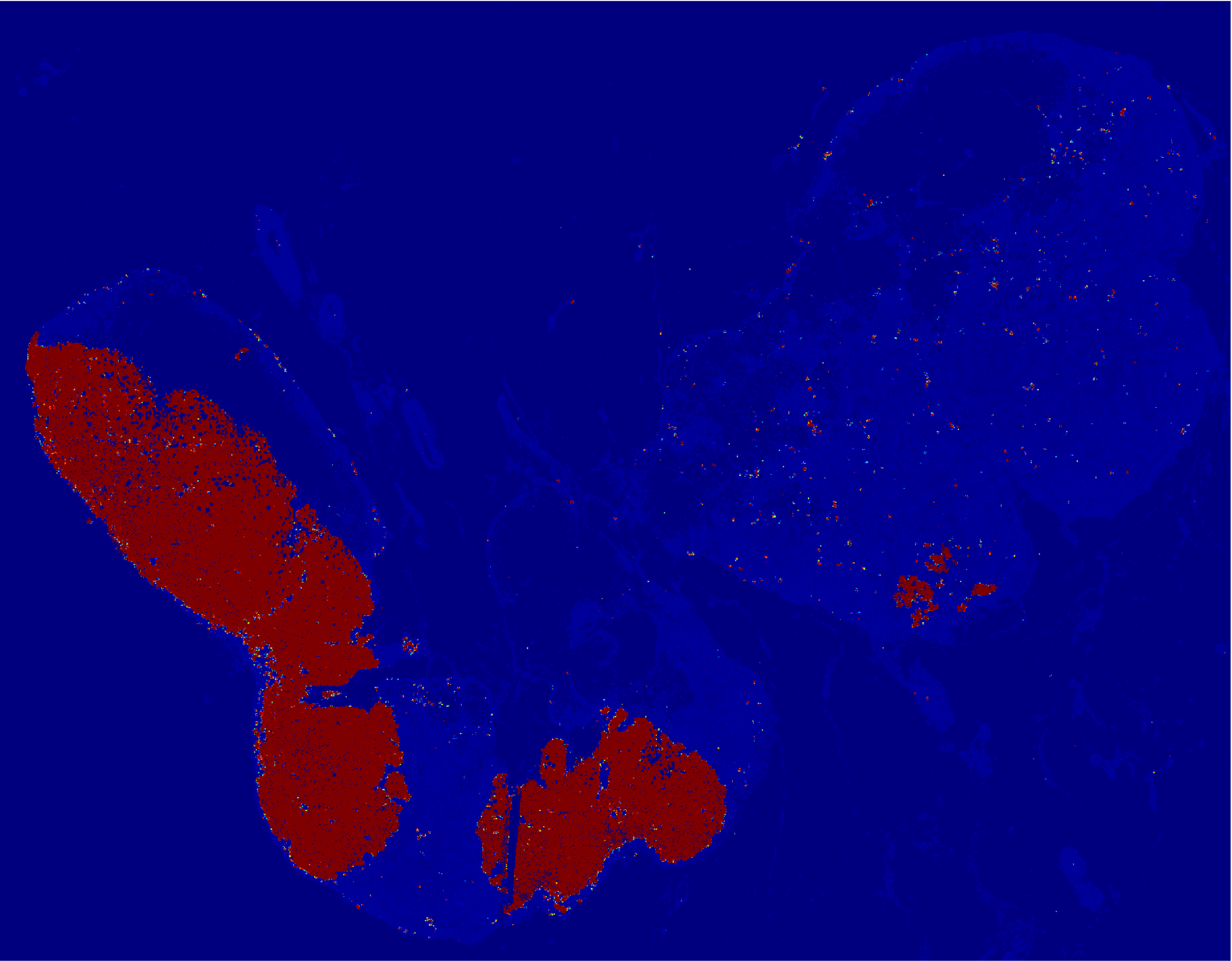}}
 		\centerline{(d) Curriculum-II}
 	\end{minipage}
     	&
 \vspace{6mm}
 	\begin{center}
     \includegraphics[scale=0.075]{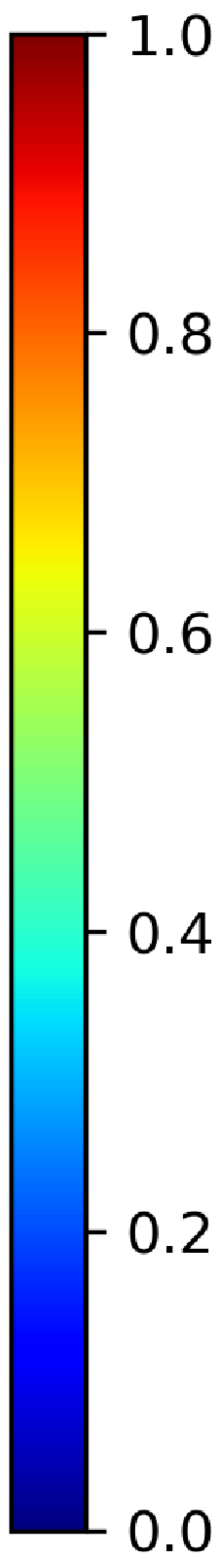}
  	\end{center}
\end{tabular}
\vspace{2mm}
\caption{Predicted tumor probability heat-maps on Camelyon16 test set with RSP (\textbf{top} row) and MoCo (\textbf{bottom} row) methods. Note. (a) Original WSI's with overlaid manual ground truth (shown in blue) depicting the region containing both macro and micro-metastases.}
\label{Fig:Predicted tumor probability heat-maps on Camelyon16 test set}
\vspace{-4mm}
\end{figure*}

\begin{figure*}
\centering
\begin{tabular}{
  @{}>{\centering\arraybackslash}m{\dimexpr.80\textwidth-\tabcolsep\relax}
  >{\centering\arraybackslash}m{\dimexpr.0001\textwidth-\tabcolsep\relax}@{}
  }
 	\begin{minipage}{0.245\linewidth}
 		\centerline{\includegraphics[height = 1.6cm, width = 3.4cm]{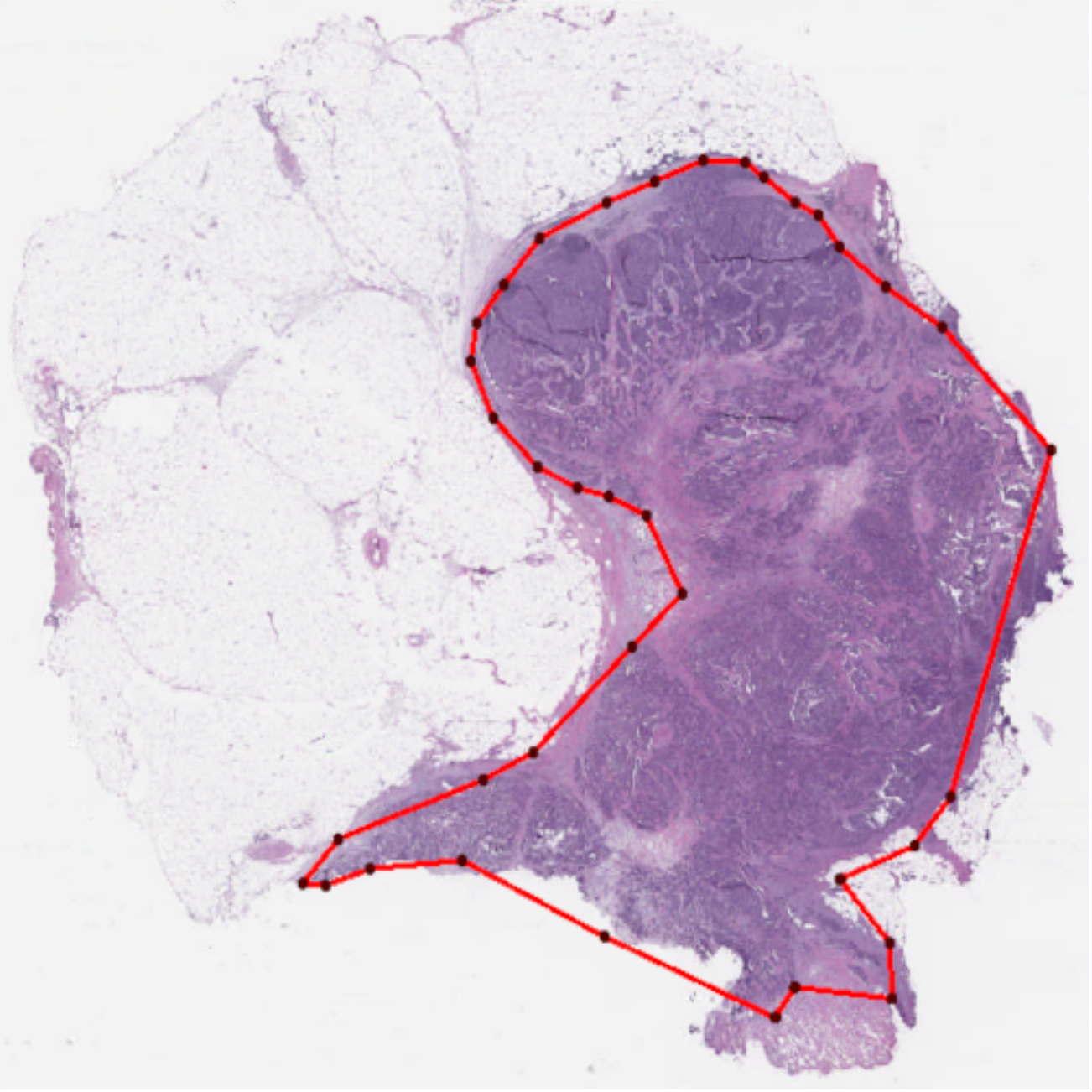}}
	\end{minipage}
    \hfill
    \begin{minipage}{0.245\linewidth}
 		\centerline{\includegraphics[height = 1.6cm, width = 3.4cm]{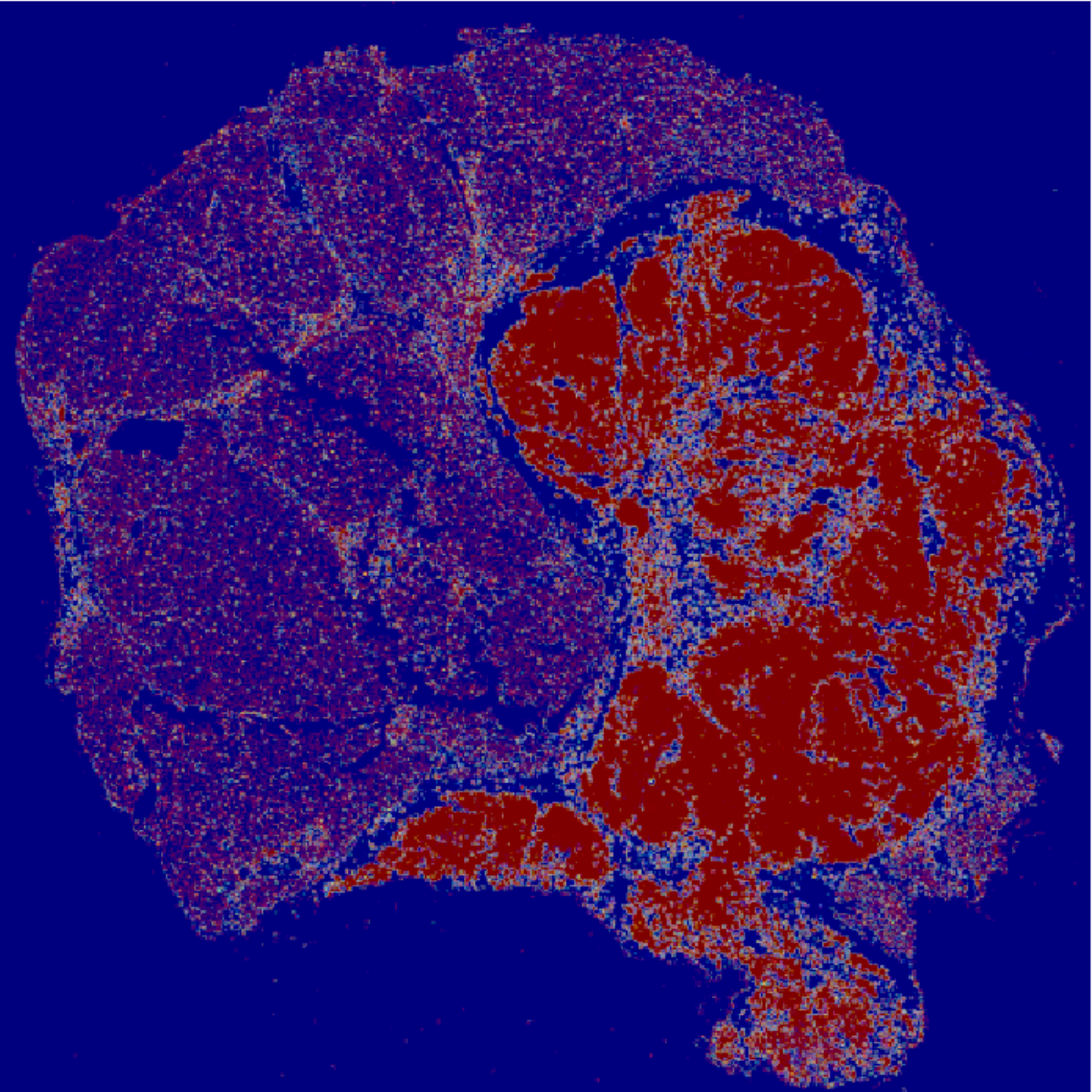}}
 	\end{minipage}
    \hfill
    \begin{minipage}{0.245\linewidth}
 		\centerline{\includegraphics[height = 1.6cm, width = 3.4cm]{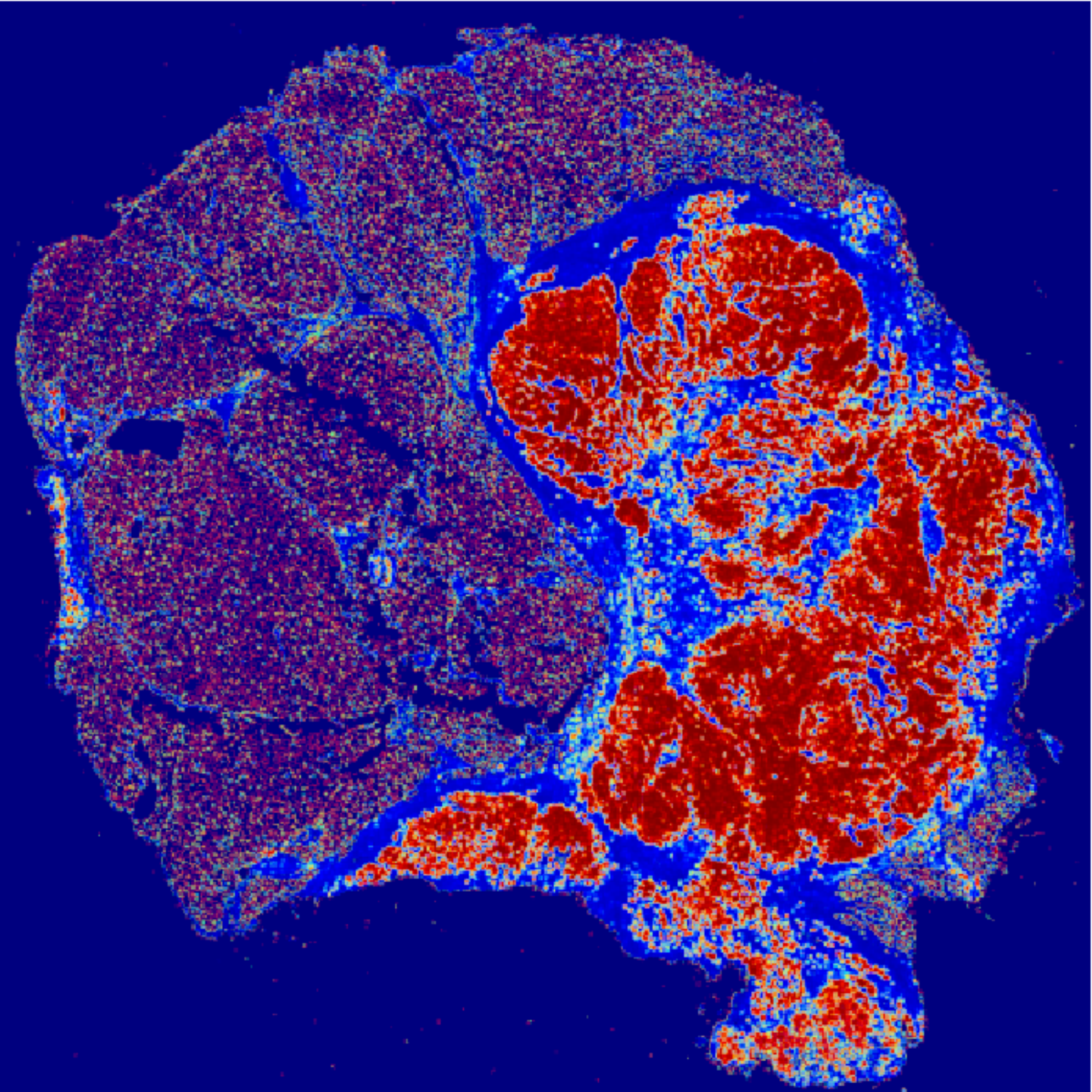}}
 	\end{minipage}
    \hfill
    \begin{minipage}{0.245\linewidth}
 		\centerline{\includegraphics[height = 1.6cm, width = 3.4cm]{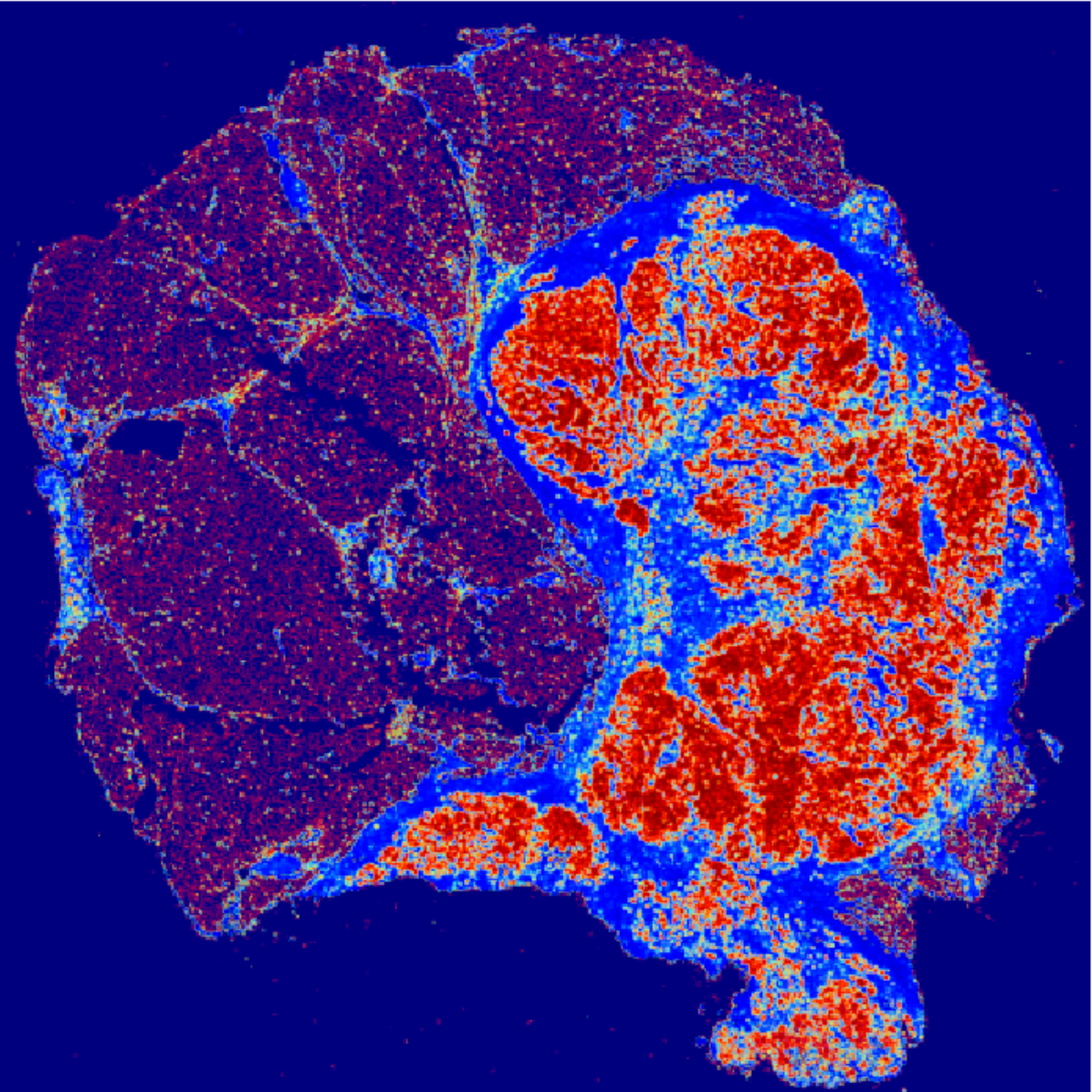}}
 	\end{minipage}
 	\vfill
    \begin{minipage}{0.245\linewidth}
 		\centerline{\includegraphics[height = 1.6cm, width = 3.4cm]{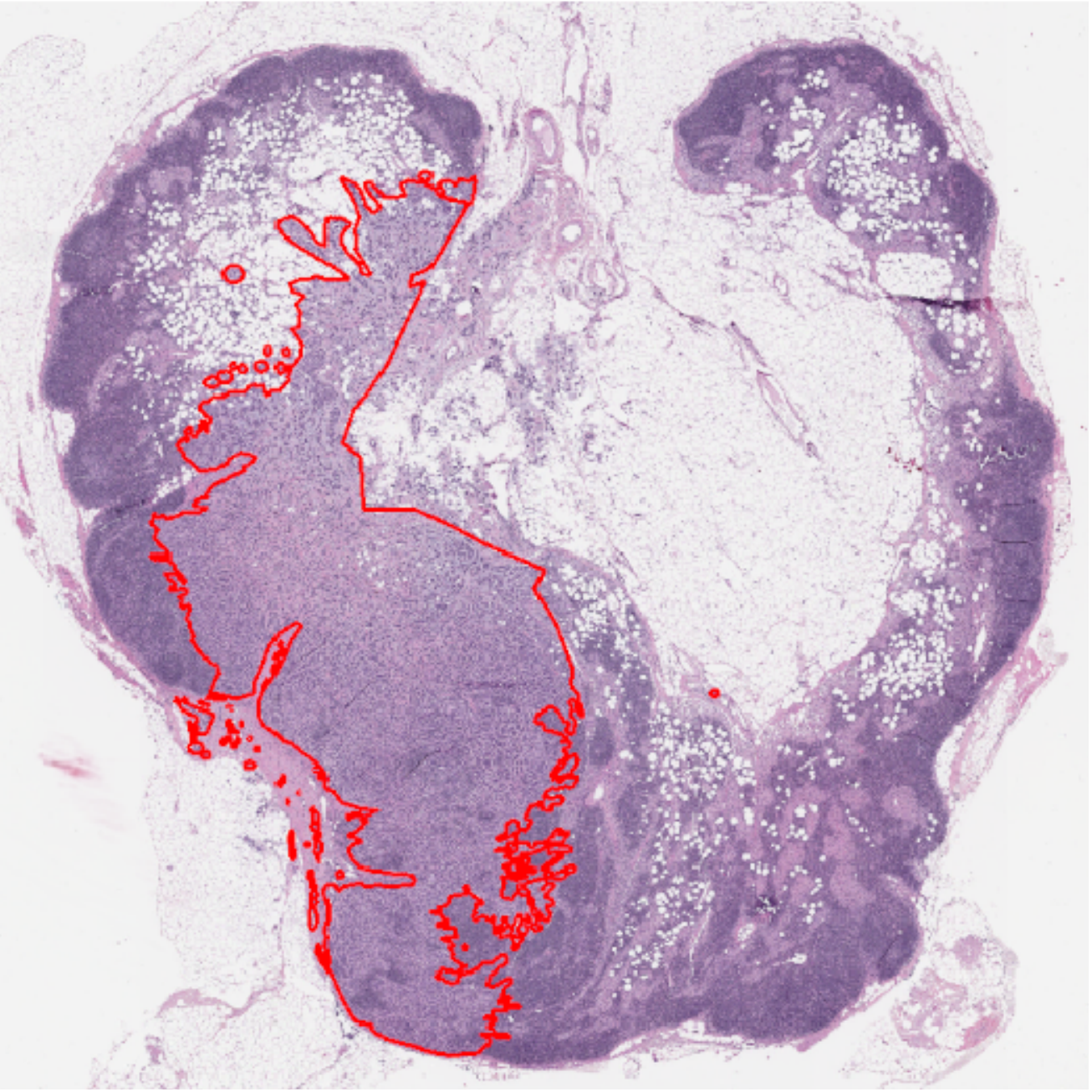}}
 		\centerline{(a) Original}
	\end{minipage}
    \hfill
    \begin{minipage}{0.245\linewidth}
 		\centerline{\includegraphics[height = 1.6cm, width = 3.4cm]{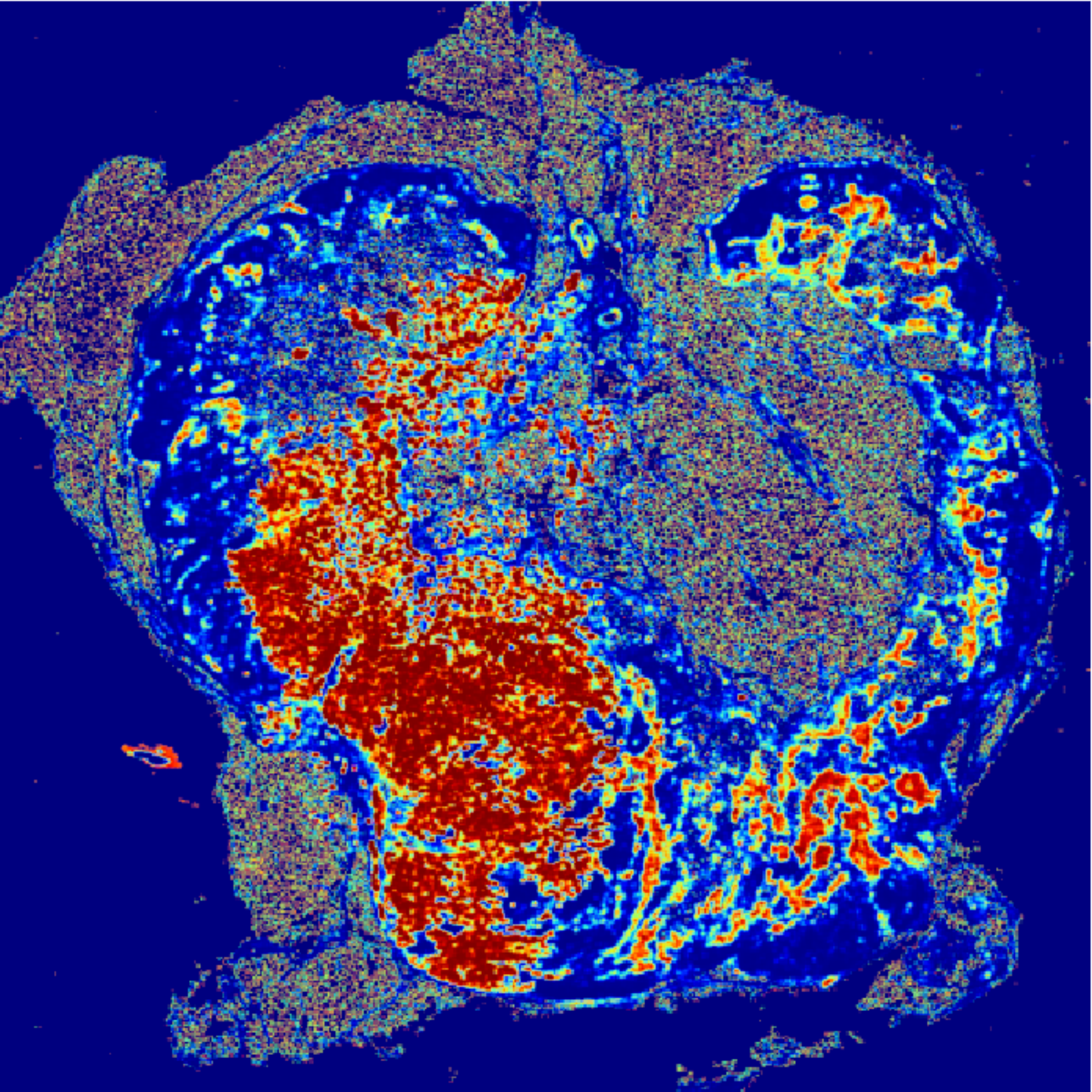}}
 		\centerline{(b) Baseline}
 	\end{minipage}
    \hfill
    \begin{minipage}{0.245\linewidth}
 		\centerline{\includegraphics[height = 1.6cm, width = 3.4cm]{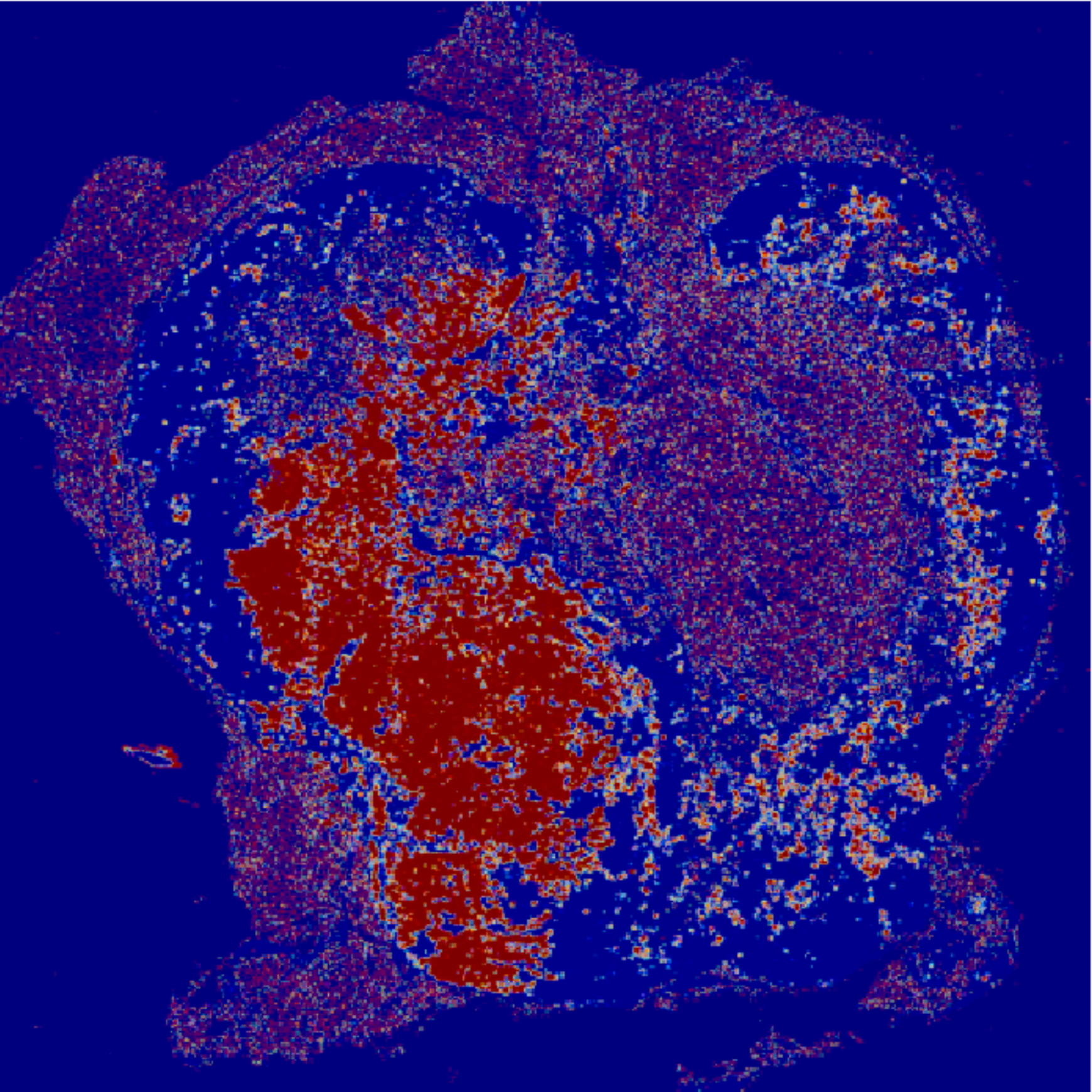}}
 		\centerline{(c) Curriculum-I}
 	\end{minipage}
    \hfill
    \begin{minipage}{0.245\linewidth}
 		\centerline{\includegraphics[height = 1.6cm, width = 3.4cm]{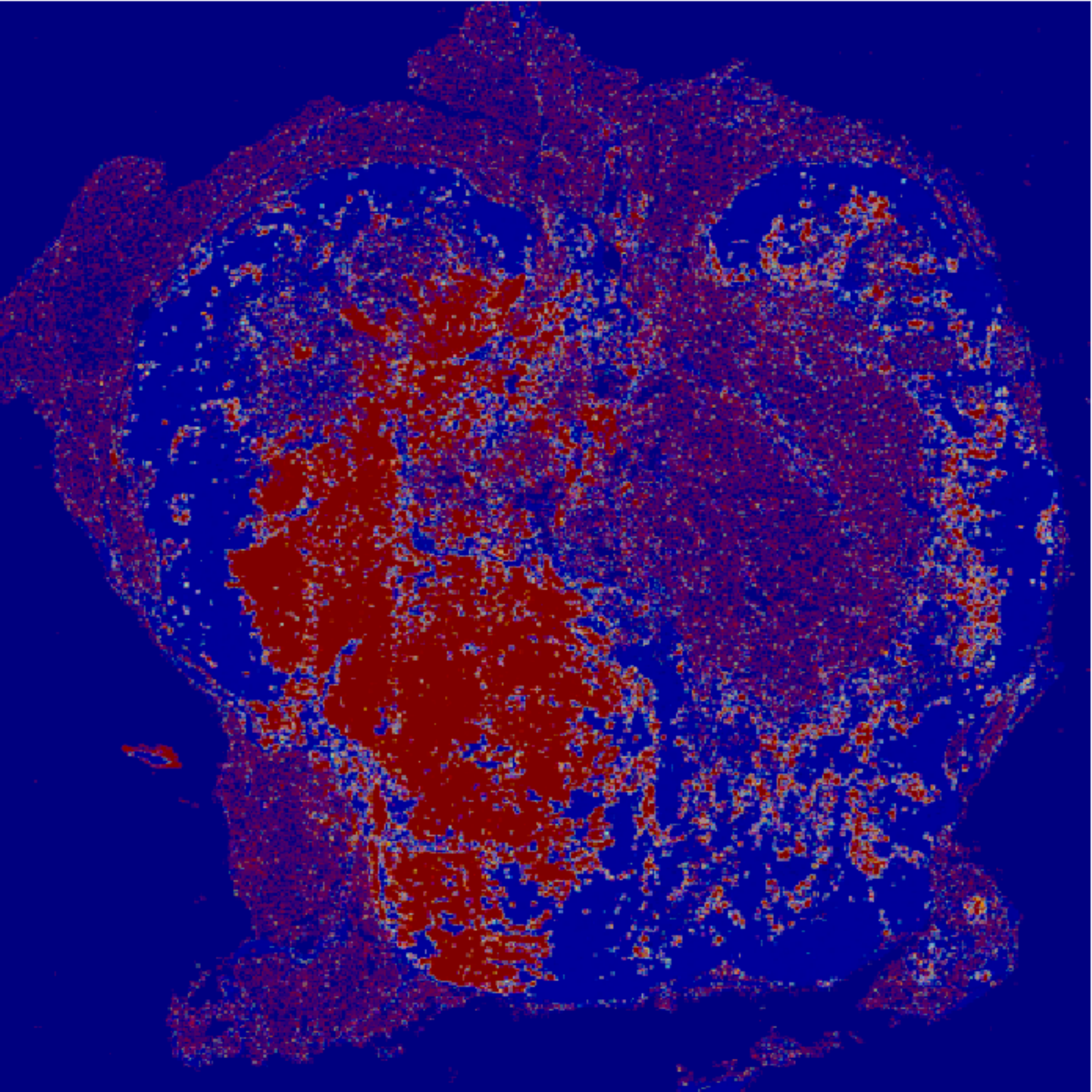}}
 		\centerline{(d) Curriculum-II}
 	\end{minipage} 
 	&
 \vspace{6mm}
 	\begin{center}
     \includegraphics[scale=0.075]{color_bar}
  	\end{center}
\end{tabular}
\vspace{2mm}
\caption{Out-of-distribution prediction results showing tumor probability heat-maps on MSK test set (\textbf{target}) trained from Camelyon16 (\textbf{source}) with RSP (\textbf{top} row) and MoCo (\textbf{bottom} row) methods. Note. (a) Original WSI's with overlaid manual ground truth (shown in red) annotated by our in-house pathologist.} 
\label{Fig:Predicted tumor probability heat-maps on MSK test set}
\vspace{-3mm}
\end{figure*}

\vspace{2mm}
\noindent
\textbf{WSI-level Classification.} The quantitative results are summarized in Table \ref{tab:overall_results} and qualitative results are shown in Figure \ref{Fig:Predicted tumor probability heat-maps on Camelyon16 test set}, \ref{Fig:Predicted tumor probability heat-maps on MSK test set}. On the Camelyon16 dataset, we achieved statistically significant improvement in accuracy (Acc) and AUC, with a minimum score of 9.3\% and 3.4\%, respectively, with Curriculum-I stage against the standard baseline; while the performance of Curriculum-II improved with a minimum Acc and AUC score of 6.2\% and 6.5\%, respectively, over the baseline, using Random and RSP pretrained methods. On the other hand, the MoCo performance improved marginally with a 1.7\% increase in AUC with Curriculum-II vs. baseline. Notably, our proposed HaDCL method achieves the best AUC score of 0.942 with 400K labeled samples compared to an AUC of 0.925 of the top-1 winning method of Camelyon16 \cite{wang2016deep}, which was trained in a fully-supervised manner with millions of image patches. 

We conducted further experiments to evaluate the effective robustness of SSL methods to out-of-distribution (OOD) data. For this, we first pretrain followed by fine-tuning the model on Camelyon16 but tested on MSK dataset. We observed significant improvement in SSL methods on OOD data, particularly when fine-tuned with curriculum-I and II approaches over the standard baseline, as shown in Table \ref{tab:overall_results}. We observed larger gains with minimum improvement in Acc and AUC score of 11.5\% and 16.7\%, respectively, with Curriculum-I stage vs. standard baseline; whereas Curriculum-II's performance further improved over baseline, with a minimum increase in Acc and AUC score of 38.4\% and 22.6\%, respectively, using Random and RSP methods. Furthermore, the performance with MoCo also improved with a 2.2\% increase in Acc and AUC score with both Curriculum-I and -II over baseline approach. This significant improvement under domain shift is of paramount importance in real clinical settings \cite{de2021residual, stacke2019closer}, where the model trained on Camelyon16 with images acquired with higher resolution ($0.25 \mu m/pixel$) can generalize satisfactorily to the OOD MSK test set, which was acquired with a lower resolution ($0.5 \mu m/pixel$). 

Overall, our results provide evidence that representations learned by SSL methods can be further enhanced and made more generalizable to out-of-domain distribution by effectively leveraging difficult examples during fine-tuning. This observation is also \textbf{consistent} with \textbf{recent studies} in \cite{andreassen2021evolution, wu2020curricula}; where the authors have shown that the effective robustness of pretraining models can be further enhanced with a more extensive and diverse set of pretraining samples followed by fine-tuning with more difficult and noisy samples. Thus, our experimental findings clearly demonstrate that the hardness-aware curriculum learning has a superior advantage over the standard fine-tuning in improving SSL methods. Further, it is interesting to explore the effect of Curriculum fine-tuning of SSL methods under a limited labeled regime, which has significant opportunities for further enhancements as shown in a recent study in \cite{srinidhi2021self}.

\vspace{2mm}
\noindent
\textbf{Patch-level Classification.} Table \ref{tab:overall_results} presents the colorectal polyps patch-wise classification results on MHIST dataset. On this task, we didn't observe any significant improvement with the HaDCL approach over the standard baseline. However, we obtained a marginal improvement in AUC of 0.895 compared with the recent CL based method \cite{wei2021learn} with AUC of 0.882. Unlike the previous method \cite{wei2021learn}, our approach doesn't depend on the annotator agreement to determine the sample hardness but rather estimates the sample hardness via knowledge transfer from a powerful pretrained SSL model. One of the main reasons for no significant improvement is because the patch-wise dataset usually does not capture all diversity of hardness that is presented in the data compare to the level of hardness that is present in the WSI. Further, most of the curated patches are often very clean and carefully hand-picked, which lacks the level of difficulty/hardness suitable for training a model. This phenomenon has also been studied in recent work \cite{ciga2021overcoming}, where the authors show evidence that injecting hard negatives samples for patch-wise classification has been shown to degrade performance, whilst the performance improves significantly for slide-level classification tasks. Notably, this phenomenon was also shown to be consistent on vision tasks \cite{wu2020curricula, weinshall2018curriculum}, where CL has shown almost no improvement on standard benchmark datasets such as CIFAR10 and CIFAR100; while, it improves only when the task is made more difficult. 

\subsubsection{Ablation Study}
\label{sssec:Ablation Study}

\begin{table}
\centering
\caption{Impact of parameter $\alpha$ which denotes the portion of hard samples in each mini-batch. These experiments were performed on the Camelyon16 validation set with RSP based SSL method \cite{srinidhi2021self} using the Curriculum-I approach.}  
\vspace{2mm}
\label{tab:ablations}
\resizebox{0.65\linewidth}{!}{%
\begin{tabular}{@{}ccccc@{}}
\toprule \toprule [1pt]
$\boldsymbol{\alpha}$ & \textbf{0.05}   & \textbf{0.10}   & \textbf{0.15}   & \textbf{0.20}   \\ \midrule
\textbf{Accuracy}       & 0.8511 & 0.8837 & 0.7441 & 0.6511 \\ \midrule
\textbf{AUC}            & 0.8920 & 0.9176 & 0.7269 & 0.6892 \\ \bottomrule \bottomrule [1pt]
\end{tabular}}
\vspace{-3mm}
\end{table}
Table \ref{tab:ablations} shows the effect of parameter $\alpha$ that selects the portion of hard samples in each mini-batch in our formulation. We observe that varying $\alpha$ to large values ($>0.10$) will lead to a large selection of easy samples, thus deteriorating the performance; on the other hand, selecting $\alpha$ to small values ($<0.10$) over exaggerates the hard samples leading to under-fitting. Thus, we empirically found $\alpha = 0.10$ as the optimal choice based on the validation performance of the Camelyon16 set that selects sufficient hard examples to balance between over-fitting to easy samples or under-fitting to hard samples. Next, we empirically chose the value of $(a, b)$ as $(0.7, 0.2)$ in a reasonable range such that the threshold $thres$ in Eq. \ref{eq:threshold} changes at uniform speed from $a+b \rightarrow b$ which is $0.9 \rightarrow 0.2$ during the gradual course of training. More intuition on selection of $(a, b)$ with respect to training dynamics of neural network is discussed in Section \ref{sec:Method} (Curriculum-I). However, note that we fixed these parameters constant across all three datasets, and we find that the above choices of $(a, b)$ are less sensitive to data distribution. 

\section{Conclusion}
\label{sec:Conclusion}
We introduce HaDCL, a method for improving self-supervised learning to both in-domain and out-of-domain distribution data, and also slide-level and patch-wise classification tasks in histopathology. By dynamically leveraging the hard examples during downstream mini-batch fine-tuning, we learn robust features that are adaptable to different domains with significant domain shifts. Our approach is more generic and adaptable to different SSL methods and does not involve any additional overhead complexity. Through experiments, we demonstrated state-of-the-art classification results on three histology benchmark datasets with a significant performance improvement on an external test set with notable domain-shift. We believe HaDCL may prove to be a useful stepping stone in generalizing the pretrained representations to various downstream tasks under a limited annotation setting. Future research will focus on extending the approach to mixed supervision to simultaneously exploit both pixel-level and image-level annotations for slide-level prediction tasks. 

{\small
\bibliographystyle{ieee_fullname}
\bibliography{ref}
}

\end{document}